\documentclass{article}

\usepackage{arxiv}

\usepackage{amssymb}
\usepackage{amsmath}

\usepackage{lineno}
\usepackage{amsmath,amssymb}
\usepackage{graphics}
\usepackage[dvipsnames]{xcolor}
\usepackage[linesnumbered,ruled,lined, vlined]{algorithm2e}
\usepackage{multirow}%
\usepackage{hyperref}
\usepackage{booktabs}
\usepackage{amsfonts}
\usepackage{graphicx, caption, subcaption}
\usepackage{array}    
\usepackage{tikz}
\usepackage[linesnumbered,ruled,vlined]{algorithm2e}

\newcommand{\TpV}{T{\it p}V} 

\newcommand{\N}{\mathbb{N}}

\newcommand{\R}{\mathbb{R}}

\newcommand{\K}{\boldsymbol{K}}
\newcommand{\D}{\boldsymbol{D}}

\newcommand{\y}{\boldsymbol{y}}
\newcommand{\x}{\boldsymbol{x}}
\newcommand{\bu}{\boldsymbol{u}}

\newcommand{\e}{\boldsymbol{e}}
\newcommand{\w}{\boldsymbol{w}}

\usepackage{accents}
\newlength{\dhatheight}

\SetCommentSty{mycommfont}

\title{An incremental algorithm for non-convex AI-enhanced medical image processing }

\author{ Elena Morotti \\
	Dept of Political and Social Sciences\\
	University of Bologna\\
}

\begin{document}
\maketitle

\begin{abstract}
Solving non-convex regularized inverse problems is challenging due to their complex optimization landscapes and multiple local minima. However, these models remain widely studied as they often yield high-quality, task-oriented solutions, particularly in medical imaging, where the goal is to enhance clinically relevant features rather than merely minimizing global error. 

We propose incDG, a hybrid framework that integrates deep learning with incremental model-based optimization to efficiently approximate the $\ell_0$-optimal solution of imaging inverse problems. Built on the Deep Guess strategy, incDG exploits a deep neural network to generate effective initializations for a non-convex variational solver, which refines the reconstruction through regularized incremental iterations. This design combines the efficiency of Artificial Intelligence (AI) tools 
with the theoretical guarantees of model-based optimization, ensuring robustness and stability.  

We validate incDG on TpV-regularized optimization tasks, demonstrating its effectiveness in medical image deblurring and tomographic reconstruction across diverse datasets, including synthetic images, brain CT slices, and chest-abdomen scans. Results show that incDG outperforms both conventional iterative solvers and deep learning-based methods, achieving superior accuracy and stability. Moreover, we confirm that training incDG without ground truth does not significantly degrade performance, making it a practical and powerful tool for solving non-convex inverse problems in imaging and beyond.  

\end{abstract}


%



\section{Introduction} \label{sec:intro}

Medical imaging plays a fundamental role in modern healthcare, enabling the visualization of anatomical structures and physiological processes for diagnostic, therapeutic, and research purposes. A wide range of imaging modalities, including X-ray computed tomography (CT), magnetic resonance imaging (MRI), ultrasound, and nuclear imaging, provide valuable insights into the human body. However, producing high-quality medical images remains a challenging task, due to the presence of noise, the scarcity of acquired data, and the stringent constraints on processing time. These challenges arise both in the reconstruction of images from raw measurements and in post-processing techniques aimed at enhancing image quality.

Let $\x^{GT}\in\R^n$ be the vectorized ground truth image and $\y \in \R^m$ the corresponding observed data. The imaging process can be modeled as:
\begin{equation}\label{eq:forward_problem}
\y = \K\x^{GT} + \e, 
\end{equation}
where  $\K \in \R^{m \times n}$, with $m\leq n$, is a linear operator describing the measurement or degradation process, and  $\e \in \R^{m}$ accounts for noise or perturbations.
This setup defines a typical inverse problem, which is often addressed through a variational formulation combining two main components: a data-fidelity term $\mathcal{F}(\x)$, enforcing agreement with the observed data $\y$, and a regularization term
$\mathcal{R}(\x)$, promoting desired properties in the reconstructed image.
Due to its reliance on an explicit mathematical description of the acquisition process and prior knowledge, this strategy is commonly referred to as Model-Based (MB).

The choice of the prior function $\mathcal{R}(\x)$ in MB approaches plays a critical role and is typically tailored to the specific imaging task.
In medical image processing, the objective is not merely to reconstruct anatomically perfect images, but to enhance features that are diagnostically relevant. Commercial software, for instance, often focuses on producing grayscale images that make subtle pathological signs—such as small lesions or low-contrast anomalies—easily detectable for clinicians.

To this end, the regularizer $\mathcal{R}$ is commonly designed to promote sparsity not on the image $\x$ itself, but on its gradient, i.e., on the vector of finite differences $\D\x \in \R^{2n}$, defined as:
\begin{equation}
\D \x = \begin{bmatrix}
\D_h \x \\ \D_v \x
\end{bmatrix}\in \R^{2n},
\end{equation}
where $\D_h$ and $\D_v$ represent discrete horizontal and vertical derivatives. Alternatively, regularization may be applied to the gradient-magnitude image $| \D \x | \in \R^n$, with entries:
\begin{equation} \label{eq:gradient_magnitude}
    \left( | \D \x | \right)_i =\sqrt{\left( \D_h \x \right)_i^2 + \left( \D_v \x \right)_i^2}, \quad \forall i =1, 2, \dots, n.
\end{equation}

Among sparsity-inducing priors, the $\ell_0$ quasi-norm, which counts the number of non-zero entries, is known to be particularly effective. 
When applied to the gradient-magnitude, it encourages reconstructions that are piecewise smooth while preserving sharp edges.
The resulting MB formulation thus reads:
\begin{equation}\label{eq:L0_OptimConstr}
\min_{\x \in \R^n} ||\ |\D\x|\ ||_0, \quad \text{subject to } \quad \y = \K\x.
\end{equation}

However, since the $\ell_0$ minimization is inherently not differentiable and computationally challenging, practical implementations often rely on continuous relaxations, such as $\ell_1$-norm regularization or nonconvex penalties approximating the $\ell_0$ norm.
In \cite{trzasko2009relaxed, montefusco2013fast}, different nonconvex families of sparsifying functions depending on a positive parameter and homotopic with the $\ell_0$ semi-norm are considered and applications of nonconvex penalties have been proposed in a great number of medical imaging applications. 
In this study, we consider the family of p-norm relaxations for $0<p\le1$ \cite{chartrand2007exact, chartrand2009fast}, and use them in the context of a Lagrangian approach, yielding a reformulation of Problem \eqref{eq:L0_OptimConstr} as the following unconstrained problem:
\begin{equation}\label{eq:optim_Rp}
    \min_{\x \in \R^n} \frac{1}{2}|| \K\x - \y^\delta ||_2^2 + \lambda\ ||\ |\D\x|\ ||_p^p,
\end{equation}
with a suitable parameter $\lambda>0$ \cite{wipf2010iterative}. 

To tackle the non-convex minimization problem in \eqref{eq:optim_Rp}, a common strategy is to minimize a convex majorizer of the original objective iteratively. This leads to Iterative Reweighted (IR) algorithms, which have been widely used and have shown improved performance over standard convex approaches.
However, IR algorithms do not guarantee convergence to the global minimum of the original non-convex problem. Moreover, they tend to be computationally demanding, as they require solving a sequence of subproblems with high precision across many iterations.

In this regard, the framework proposed in \cite{lazzaro2019nonconvex} is particularly relevant, as it explicitly addresses the challenge of converging to a meaningful local minimum in non-convex optimization. Their incremental scheme progressively updates the model parameters across iterations, effectively guiding the algorithm toward highly sparse solutions in the gradient domain.
Our work builds upon the same core philosophy, adopting this incremental strategy as a foundation and automatic parameter updating rules that are appealing for medical applications, where no heuristic fine-tuning (based on many trials) is possible. However, we diverge in two key aspects: we employ a different non-convex regularizer (the $\ell_p$-norm) and we use a different inner optimization method, relying on the primal-dual Chambolle–Pock algorithm rather than a forward–backward scheme.

\begin{figure}
    \centering
    \includegraphics[width=0.85\linewidth]{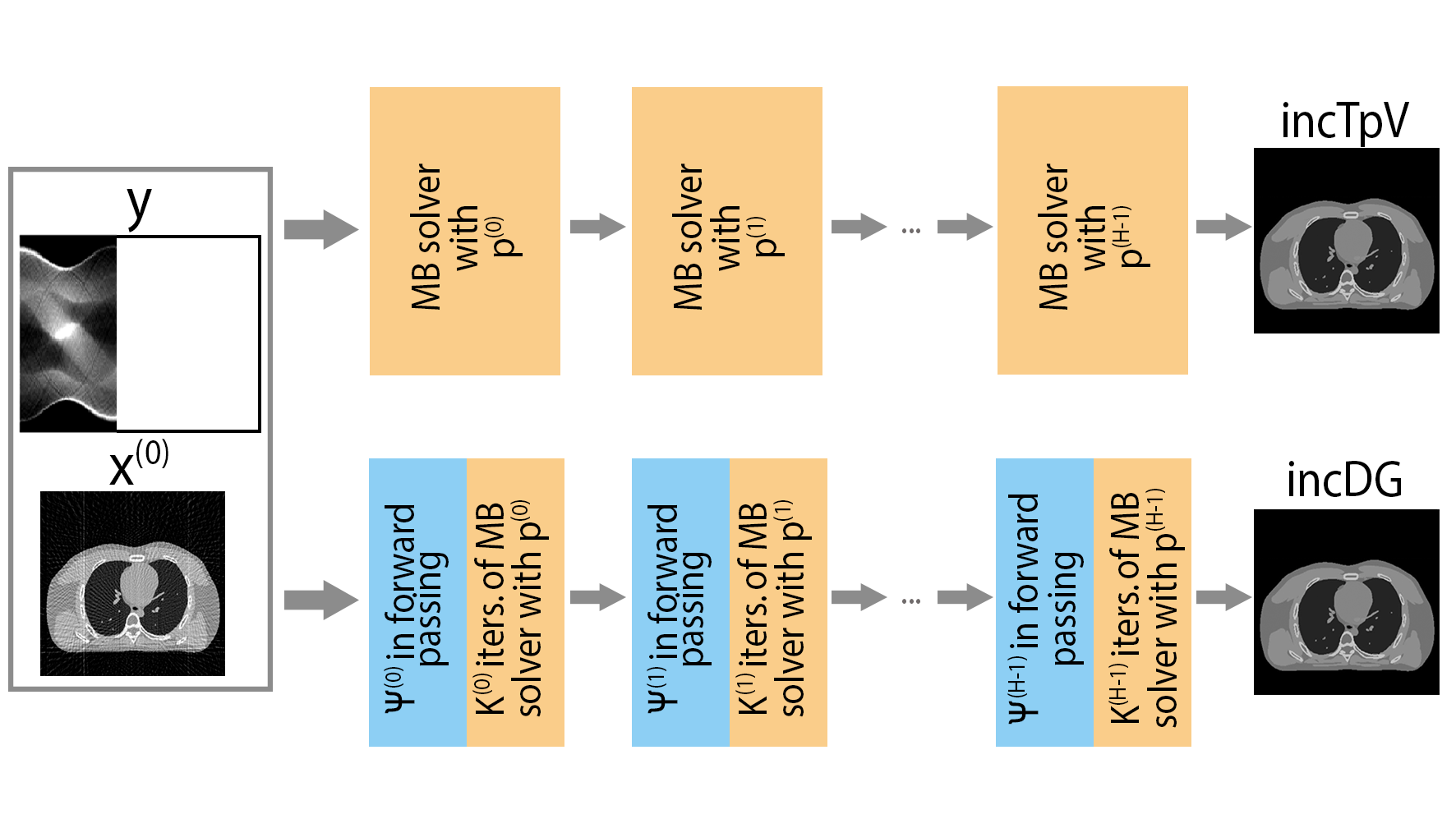}
    \caption{Sketch of the considered incremental approaches, applied to tomographic image reconstruction. }
    \label{fig:GraphicalAbstract}
\end{figure}

To further accelerate the incremental scheme, this work integrates state-of-the-art Artificial Intelligence (AI) tools. 
Specifically, we embed an end-to-end convolutional neural network to support the solution of each reweighted subproblem, following the Deep Guess (DG) strategy recently introduced in \cite{LoliPiccolomini2025DeepGuess}. At each iteration, the network produces a refined estimate by mapping the current solution closer to the ground truth, then the subproblem solution is fine-tuned by a few steps of the MB algorithm. This hybrid method, referred to as incDG throughout the manuscript, combines both the structure of the incremental variational model and the efficiency of deep learning-based acceleration. A schematic comparison with the fully MB baseline (denoted as incTpV) is provided in Figure \ref{fig:GraphicalAbstract}.

To some extent, while significantly accelerating running times via the fast forward pass of pre-trained networks, this hybrid scheme preserves the interpretability and a mathematical characterization of the final solution by the model-based approach.\\

This work contributes to the field of non-convex regularized inverse problems and model-based imaging, as summarized in the following.
\begin{itemize}
    \item Extension of incremental non-convex optimization strategies.\\
    Building on the incremental scheme introduced in \cite{lazzaro2019nonconvex}, we adopt a different non-convex prior (a smoothed $\ell_p$ norm on the image gradient magnitude), leading to different weighted subproblems, and use the efficient Chambolle–Pock algorithm. 
    This results in a flexible and robust framework tailored to strongly non-convex settings, with improved sparsification in the gradient domain.
    \item Hybrid integration of AI and non-convex variational models.\\
    Rather than replacing the optimization process, the network guides and accelerates it by producing effective initial guesses. This hybrid approach inherits the interpretability and mathematical grounding of variational methods, while benefiting from the computational efficiency and speed of deep learning.
    \item Robust performance in realistic and ground-truth-free scenarios.\\
    The proposed method is validated on challenging medical imaging problems, including deblurring and CT reconstruction from subsampled data. Notably, it achieves high-quality and contrast-enhanced results even when trained without ground truth data, demonstrating both the method's practical applicability and intrinsic stability.
\end{itemize}

The organization of the paper is as follows. 
Section \ref{sec:model} introduces the model-based incremental approach, providing detailed derivations of the model reformulations and the underlying algorithms. 
Section \ref{sec:proposed} presents the acceleration strategy based on deep learning, including a description of the neural network architecture and training process.
Experimental results validating the proposed method are reported and discussed in Section \ref{sec:results}. Finally, concluding remarks are provided in Section \ref{sec:concl}.

\section{Incremental model-based approach for non-convex optimization} \label{sec:model}

In this section, the model-based incremental approach is derived, starting from the formulation of the non-convex optimization problem, then introducing the iterative reweighted strategy to handle the non-convex regularizer, and finally embedding it within an incremental scheme that progressively updates the model parameters and the solution.

\subsection{Preliminaries on non-convex optimization}\label{ssec:nonconvex}
In most imaging applications, the image degradation model can be formulated as the linear forward problem \eqref{eq:forward_problem}, and addressing the ill-posedness of the associated inverse problem through advanced algorithms and processing methods is crucial. 
For this reason, a vast body of literature has proposed regularized approaches, as described in Equations \eqref{eq:L0_OptimConstr} and \eqref{eq:optim_Rp}.

\begin{figure}
    \centering
    \includegraphics[width=0.5\linewidth]{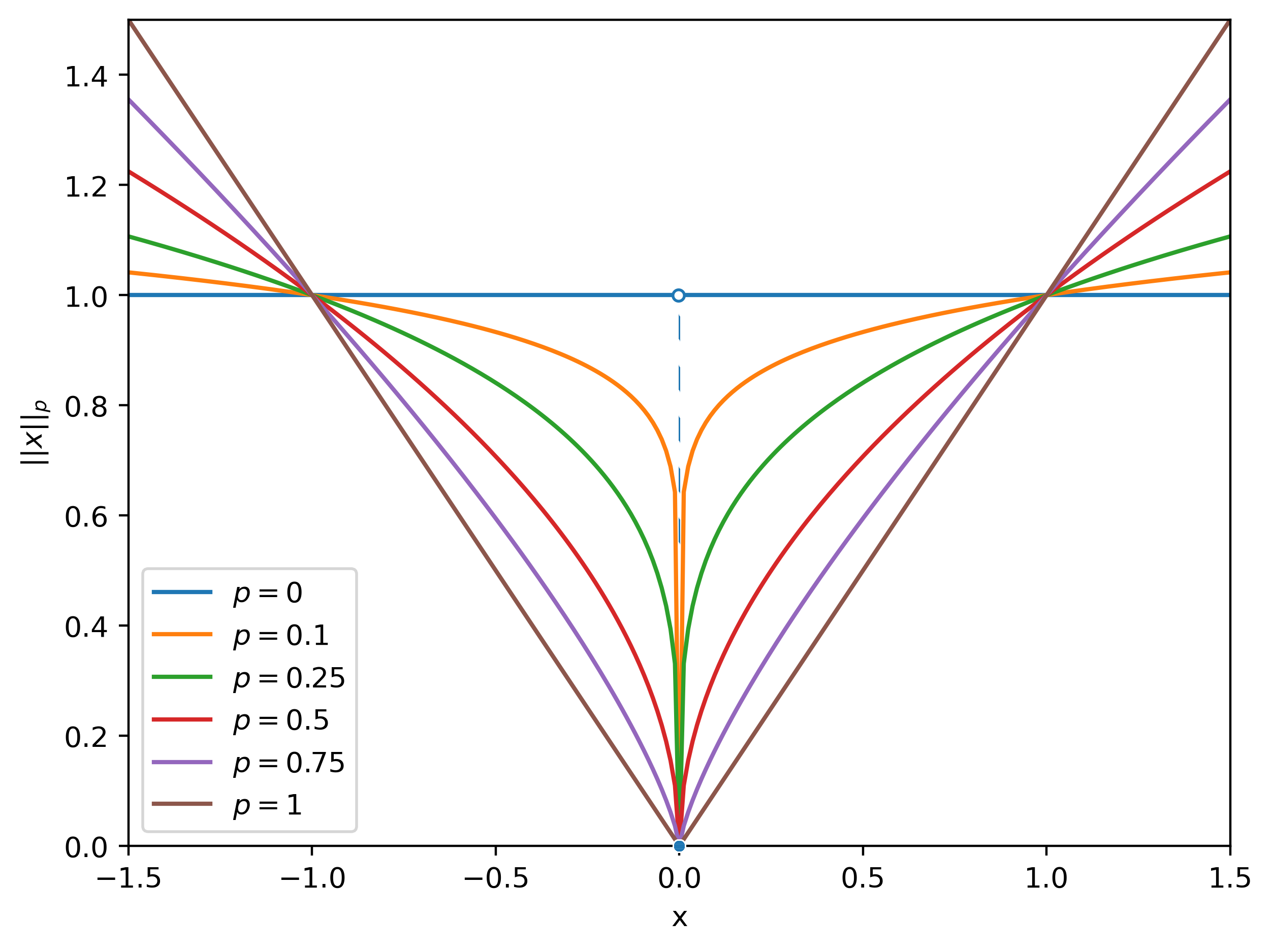}
    \caption{Plots of the $p$-norm functions $||x||_p$ for different values of $p$. The curves correspond to different values of $p$: $p=0$ (blue), $p=1$ (brown), and intermediate values approaching $p\to 0$. They illustrate the loss of convexity for $p<1$ and the non-differentiability characteristic of the $\ell_0$.}
    \label{fig:plot_pnorm}
\end{figure}

As anticipated, setting the prior function $\mathcal{R}(\x)$ with the $\ell_0$ quasi-norm of the gradient image should be optimal to preserve sharp edges and discontinuities which make the final solution effective for enhancing the detectability of diagnostically relevant structures while suppressing noise and artifacts.
Unfortunately, the $\ell_0$ quasi-norm is inherently non-continuous and non-smooth, making differentiation impossible in the conventional sense, and its direct minimization leads to a combinatorial problem, which is NP-hard.
Several algorithms have been developed to efficiently solve the problem in Equation \eqref{eq:L0_OptimConstr} (see \cite{bioucas2007new, figueiredo2007gradient, bredies2008iterated, montefusco2010fast} and references therein), but more recent studies \cite{candes2008enhancing, mohimani2008fast} have shown that using alternative non-convex sparsity-inducing functions in a constrained framework like \eqref{eq:L0_OptimConstr} can enable even more accurate and efficient recovery of sparse signals. 
Examples of these non-convex functions are the $\ell_p$ norms for $0<p<1$ \cite{chartrand2009fast, mohimani2010sparse}, which replicate the sparsity-promoting behavior of the $\ell_0$ quasi-norm more closely than the $\ell_1$ norm, as visible in Figure \ref{fig:plot_pnorm}.
Further examples are the {\it log-sum}, the {\it log-exp} and the {\it atan} sparsity inducing functions \cite{candes2008enhancing, montefusco2013fast, mourad2010minimizing}. \\
In this study, we impose sparsity to the gradient-magnitude image through the $\ell_p$ norm, thus we reformulate the original problem \eqref{eq:L0_OptimConstr} as:
\begin{equation}\label{eq:Lp_OptimConstr}
    \min_{\x} ||\ |\D\x|\ ||_p^p \qquad s.t.\quad  \y = \K\x.
\end{equation}
In this case, the prior is typically called Total $p$-norm Variation function and denoted as $\TpV(\x) := ||\ |\D\x|\ ||_p^p$. It is defined for a suitable choice of $p \in (0, 1]$, playing the role of {\it sparsity} parameter.

It is worth noting that the regularizaion function becomes the widely used and convex Total Variation when $p=1$.  In this case, also the unconstrained model stated as in \eqref{eq:optim_Rp} is convex and its global minimum can be computed using iterative algorithms such as Scaled Gradient Projection \cite{bonettini2008scaled, piccolomini2016fast} or Chambolle-Pock (CP) \cite{chambolle2011first}.

\subsection{Iterative Reweighted algorithm for TpV minimization \label{ssec:IR_CP}}
To address the non-convex minimization problem in \eqref{eq:Lp_OptimConstr} for a fixed $p \in (0, 1)$, Iterative Reweighting (IR) strategies have been proposed that minimize convex $\ell_1$ majorizations of the original regularizer \cite{candes2008enhancing, daubechies2010iteratively}. 
More precisely,  as the $\TpV$ prior  fully preserves the separability properties typical of both $\ell_0$ and $\ell_1$ norms, calling $\bu = |\D\x| \in \R^n$, our prior can be seen as:
$$
||\bu||_p^p = \sum_{i=1}^n \psi(|\bu_i|),
$$
where $\psi: \R^+ \to \R^+$ retains key properties of the $\ell_1$ norm, including continuity and differentiability.
Indeed, in our case, we can define:
\begin{equation}\label{eq:psi}
    \psi(|\bu_i|) := |\bu_i|^p \quad \forall u\in\R\symbol{92} \{0\}, 
\end{equation}
with its first derivative:
\begin{equation}\label{eq:psiprime}.
    \psi\prime(u) = \frac{p}{|u|^{1-p} } \quad \forall u>0 
\end{equation}
This function enjoys the following important properties:
\begin{itemize}
    \item $\psi(u)$ is concave and non-decreasing for $u\in (0, \infty)$;
    \item $\psi(u)$ is singular in $u=0$;
    \item its derivative $\psi^{\prime}(u)$ satisfies:     
    \[
        \begin{cases}
            \psi^{\prime}(u)>0 & \text{if } u> 0 \\
            \psi^{\prime}(u)\gg 0 & \text{as } u\to 0^+
        \end{cases}
    \]
\end{itemize} 

Now, the non-convex $\TpV$ objective in \eqref{eq:Lp_OptimConstr} is replaced by its convex, easy to minimize upper bound $R(\x)$ defined by a {\it local} linear approximation of $\mathcal{R}$ in proximity to the point {$\bar{\bu} = \bar{|\D\x|} \in\R^n$} \cite{zou2008one}:
\begin{equation}
     R(\x) = \sum_{i=1}^n \Psi(|\D\x|_i) := \sum_{i=1}^n \psi(|\bar{\bu_i}|) + \psi^{\prime}(|\bar{\bu_i}|)(|\bu_i|-|\bar{\bu_i}|)).
\end{equation}
The majorization property $R(\x) \ge \mathcal{R}(\x)$ is easily obtained.
Additionally, as the original constrained minimization problem described in  \eqref{eq:Lp_OptimConstr} can be reformulated as a convex unconstrained problem presented in \eqref{eq:optim_Rp}, we incorporate the local linear approximation into the construction of the augmented Lagrangian functional. 
We thus get:
\begin{equation}\label{eq:passaggiIR}
\begin{split}
     \min_{\x} \| \K\x - \y \|_2^2  + \lambda R(\x) &  = 
     \min_{\x} \| \K\x - \y \|_2^2  + \lambda \sum_{i=1}^n \psi(|\bar{\bu_i}|) + \psi^{\prime}(|\bar{\bu_i}|)(|\bu_i|-|\bar{\bu_i}|)) \\
    &  = \min_{\x} \| \K\x - \y \|_2^2  + \lambda \sum_{i=1}^n  \psi'(|\bar{\bu_i}|)(|\bu_i|)\\
    &  = \min_{\x} \| \K\x - \y \|_2^2  + \lambda \sum_{i=1}^n  \frac{p}{|\D\bar{\x}|_i^{1-p}} |\D\x|_i.
\end{split}
\end{equation}
Deriving from the reformulation in \eqref{eq:passaggiIR} for local approximations, the IR model iteratively addresses the following minimization \cite{ochs2015iteratively, sidky2014cttpv}:
\begin{equation}\label{eq:inverse_problem_weighted}
     \x^{(k+1)} = \arg\min_{\x\in\Omega} \| \K\x - \y \|_2^2  + \lambda\cdot || \ \w^{(k)} \odot | \D\x| \ ||_1,
\end{equation}
where $\odot$ represents the element-wise product and the weights $\w^{(k)} \in \R^n$ depend on the current iterate $\x^{(k)}$ and on the sparsity parameter $p$, being defined as:
\begin{equation}
    \w^{(k)}_i := \left( \w(\x^{(k)}, p) \right)_i = \frac{p}{|\D\x^{(k)}|_i^{1-p} + \xi} \quad \forall i=1, \dots, n
\end{equation}
with $\xi>0$ is a predefined smoothing parameter.
In \eqref{eq:inverse_problem_weighted} we also enrich the model with a non-negative constraint onto the image entries $\x_i$, by denoting with $\Omega := \{\x \in \R^n; \x_i \geq \boldsymbol{0} \ \forall i=1, \dots n\}$ the feasible set of medical radiological images \cite{montefusco2013fast}. 

Inspired by \cite{sidky2014cttpv, loli2021model}, we solve each  convex $\ell_1$ subproblem \eqref{eq:inverse_problem_weighted} with the popular Chambolle-Pock (CP) algorithm, which implements a primal-dual approach to tackle convex minimizations. It has been introduced in this original formulation in \cite{chambolle2011first}, to solve an unconstrained minimization given by:
\begin{equation*}
    \min_{\x \in \R^n} F_1(M\x) + F_2(\x),
\end{equation*}
where both $F_1$ and $F_2$ are real-valued, proper, convex, lower semi-continuous functions, and $M$ is a linear operator from $\R^n$ to $\R^s$. 
As there are no constraints on the smoothness of either $F_1$ and $F_2$, the CP algorithm allows for rapid prototyping of optimization problems, fitting the variety of those arising in image processing.
To fit our model \eqref{eq:inverse_problem_weighted}, it is enough to set:
\begin{equation*}
    \begin{cases}
        F_1(M\x) = \frac{1}{2} || \K\x - \y ||_2^2 + \lambda\cdot || \ \w^{(k)} \odot | \D\x| \ ||_1,\\
        F_2(\x) = \iota_{\Omega}(\x),        
    \end{cases}    
\end{equation*}
where $\iota_{\Omega}(\x)$ represents the indicator function of our feasible set $\Omega$, and the linear operator $M \in \R^{s \times n}$ is built by the row-wise concatenation of $\K$ and $\D$, such that $M = \left[ \K; \D \right]$. \\
In the Iterative Reweighted strategy, each iteration requires solving a convex $\ell_1$ subproblem \eqref{eq:inverse_problem_weighted} exactly. 
However, in practice, each subproblem \eqref{eq:inverse_problem_weighted} is solved approximately, by running a limited number $\texttt{k}^{CP}$ of iterations of the CP algorithm.  \\
For deeper details about the CP iterations, the reader can refer to \cite{sidky2012convex, LoliPiccolomini2025DeepGuess}.

At last, we remark that the IR algorithm needs a starting iterate $\x^{(0)}$ as input, whereas it is stopped when a numerical convergence is approached, defined by:
$$
\frac{|| \x^{(k)} - \x^{(k-1)} ||_2}{|| \x^{(k-1)} ||_2 + 10^{-6}} < \tau_x  \quad \mbox{and} \quad
\frac{|| \K\x^{(k)} - \y ||_2}{\sqrt{m}||\y||_\infty} < \tau_\mathcal{F},
$$
with suitable parameters $\tau_x$ and $\tau_\mathcal{F}$, or when the maximum number $\texttt{k}^{IR}$ of iterations have been performed.

The resulting IR algorithm solving the non-convex problem \eqref{eq:Lp_OptimConstr} is summarized in Algorithm \ref{alg:cp_tpv}.

\begin{algorithm}[hbt!]
\caption{Iterative Reweighted algorithms for $\TpV$-regularized model ($\TpV$)}\label{alg:cp_tpv}
{\bf input:} $\K, \y, \tilde{\x}, \lambda, p, \texttt{k}^{CP}, \texttt{k}^{IR}, \tau_x, \tau_\mathcal{F}$\;
\tcp{Initializations}
$\x^{(0)} = \tilde{\x}$; \\
$\xi=2\cdot 10^{-3}$;\\
k=0; count=0;\\
\Repeat{$\left(
\frac{\| \x^{(k)} - \x^{(k-1)} \|_2}{\| \x^{(k-1)} \|_2 + 10^{-6}} < \tau_x 
\ \text{and} \
\frac{\| \K\x^{(k)} - \y \|_2}{\sqrt{m}\|\y\|_\infty} < \tau_\mathcal{F}
\right)
\text{ or }
\mathrm{count} < \texttt{k}^{IR}$}{
    \tcp{Update weights}
    $\w = \frac{p}{|\D\x^{(k)}|^{1-p} + \xi} $; \\
    \tcp{Solve subproblem} 
    Compute $\x^{(k+1)} = \arg\min_{\x\in\Omega} \| \K\x^{(k)} - \y \|_2^2  + \lambda || \ \w \odot | \D\x| \ ||_1$, with ${\tt k}^{CP}$  \mbox{iterations of Chambolle-Pock algorithm};\\
    {count = count+$\texttt{k}^{CP}$};\\
    k=k+1;
} 
\Return $\x^{(k)}$
\end{algorithm}

\subsection{Incremental approach}\label{ssec:incTpV}
It is known that the computed solution of the non-convex problem \eqref{eq:Lp_OptimConstr} can deviate significantly from the desired solution approximating the ground truth $\x^{GT}$ image, due to the existence of many local minima and the usage of prefixed $p>0$.
In this regard, the choice of the initial guess $\x^{(0)}$ and of the model parameters plays a crucial role.
Whereas $\x^{(0)}$ is often computed as the solution of the $\ell_1$-regularized problem, setting (a priori) $\lambda$ and $p$ is more challenging.

In \cite{lazzaro2019nonconvex}, the authors propose an incremental scheme with automatic updating rules for the selection of both the regularization and the sparsity parameters, tuned to gradually converge to a good local minimum approaching the $\ell_0$ solution.
They use a family of priors different than $\TpV$, thus we modify that scheme to fit our $\TpV$-regularized approach. 

Specifically, we refer from now on to each incremental step with index $h$, and we compute a sequence of incremental solutions $\bar{\x}^{(h+1)}$ of a non-convex subproblem 
where both the regularization parameter $\lambda^{(h)}$ and the sparsity parameter $p^{(h)}$ depend on $h$. 

Firstly, focusing on the regularization parameter, we define a decreasing sequence $\{\lambda^{(h)}\}$ meeting the need of a lower contribution by the regularization term when approaching the solution.
Thus, given a starting value $\lambda^{(0)}>0$ and computing $\lambda^{(1)} = \frac{\lambda^{(0)}}{2}$,
we adjust the value of the subsequent regularization parameters $\lambda^{(h)}$ on the basis of the behavior of the objective function, with the strategy originally suggested in \cite{montefusco2011iterative}.
Denoting with $f^{(h)}$ 
the value of the objective function in the minimum $\bar{\x}^{(h)}$, the adaptive rule thus reads:
\begin{equation}\label{eq:UpdateLambda}
     \lambda^{(h+1)} = \lambda^{(h)} \frac{f^{(h)}}{f^{(h-1)}} \quad \forall j=1, 2, \dots.
\end{equation}

Secondly, setting an initial $p^{(0)}=1$, we update the sparsity parameter according to the following rule:
\begin{equation}\label{eq:UpdateP}
        p^{(h+1)} =  p^{(h)} \cdot \alpha_p  \quad \forall j=0, 1, \dots,
\end{equation}
where a suitable $\alpha_p$, with $0<\alpha_p<1$, makes $p$ reduce during the iterations, to progressively approach the $\ell_0$ quasi-norm.\\

With a slightly abuse of notation, combining the two updating rules \eqref{eq:UpdateLambda} and \eqref{eq:UpdateP} into our IR model yields to the resolution of:
\begin{equation}\label{eq:outer_step}
\bar{\x}^{(h+1)} \in \arg\min_{\x \in \Omega} \frac{1}{2}|| \K\x - \y ||_2^2 + \lambda^{(h)}\cdot ||\ |\D \x|\ ||_{p^{(h)}}^{p^{(h)}},
\end{equation}
at each incremental iteration of the algorithm. 
We label the resulting algorithm \emph{inc$\TpV$} in the following.
Its pseudocode is reported in Algorithm \ref{alg:inc_tpv}.

We use the Chambolle-Pock solver presented in Section \ref{ssec:IR_CP} to solve each $h$-th subproblem, and we adopt the warm starting strategy to initialize the guess $\tilde{\x}$ at each IR execution.
In practice, to implement the entire framework, we also set a priori the number $H$ of outer iterations. In addition, in each calling of the IR algorithm, we must pass the maximum number of iterations to perform as input parameter. We thus need to set a scheduler $K \in \N^H$ whose values $K_h$ correspond to the number ${\tt k}^{IR}$ of iterations allowed in the $h$-th incremental iteration, for $h=0, \dots, H-1$. 
The inc$\TpV$ algorithm thus executes a maximum of $\bar{K} = \sum_h K_h$ iterations.

\begin{algorithm}[hbt!]
\caption{Incremental algorithm for $\TpV$ minimization (inc$TpV$)}\label{alg:inc_tpv}
{\bf input:} 
$\K, \y, \lambda^{(0)}, \tilde{\x}, \alpha_p \in (0, 1), H \in \mathbb{N}, K \in \mathbb{N}^H, \tau_x, \tau_\mathcal{F}$; 
\tcp{Initializations}
$p^{(0)} = 1$;
$\bar{x}^{(0)} = \tilde{\x}$; \\
$ f^{(0)} = \frac{1}{2} || \K\bar{x}^{(0)} - \y ||_2^2 + \lambda^{(0)}\cdot || \ | \D\bar{x}^{(0)} | \> ||_{p^{(0)}}^{p^{(0)}}$;\\
$h=0$; \\
\Repeat{$h < H$}{  
    \tcp{Solve subproblem}
     Compute $\bar{\x}^{(h+1)} \in \arg\min_ {\x \in \Omega} \frac{1}{2}|| \K\x - \y ||_2^2 + \lambda^{(h)}\cdot ||\ |\D \x|\ ||_{p^{(h)}}^{p^{(h)}}$     by Algorithm \ref{alg:cp_tpv} with starting guess $\tilde{\x}$, tollerances $\tau_x$ and $\tau_\mathcal{F}$, and {\tt k}$^{IR}=K_h$\; 
    \tcp{Update parameters}
     $ f^{(h)} = \frac{1}{2} || \K\bar{\x}^{(h+1)} - \y ||_2^2 + \lambda^{(h)}\cdot || \ | \D\bar{\x}^{(h+1)} | \> ||_{p^{(h)}}^{p^{(h)}}$\;
    $p^{(h+1)} = p^{(h)} \cdot \alpha_p$\;
   {\bf if\ } $h=0${\bf :}   $\lambda^{(h+1)} = \lambda^{(0)} / 2$\;
    {\bf else\ } $\lambda^{(h+1)} = \lambda^{(h)} \cdot \frac{f^{(h)}}{f^{(h-1)}}$\; 
    \tcp{Warm starting}
    $\tilde{\x} = \bar{\x}^{(h+1)}$\;  
} 
\Return $\bar{\x}^{(H)}$\;
\end{algorithm}

\subsection{About the convergence of the incremental approach}

In \cite{lazzaro2019nonconvex}, it is proven that using a fixed smoothing parameter and a decreasing sequence of regularization parameters, as the one generated by Equation \eqref{eq:UpdateLambda},  ensures the convergence of the penalized solution to that of the original problem. 
The proof is based on the Kurdyka–Lojasiewicz property \cite{kurdyka1998gradients}, which means, roughly speaking, that the objective functions under consideration are sharp up to a parametrization.
The objective function in our model, combining a least squares data-fit term with a non-convex $\TpV$ regularizer, satisfies the Kurdyka–Lojasiewicz property because both components are semi-algebraic: the squared $\ell_2$-norm is analytic, and the non-convex regularizer is definable in an o-minimal structure. This ensures the entire functional inherits the property \cite{attouch2013convergence}.

Using a sequence of regularization parameters to solve each $\TpV$ problem (at varying values of $p$) in the incremental scheme makes the entire resolution computationally demanding, due to the number of nested inner subproblems that should be solved exactly with iterative procedures.
For this reason, Lazzaro et al., in  \cite{lazzaro2019nonconvex}, have proposed to inexactly solve each outer step, by simultaneously updating both the smoothing parameter and the regularization one.
Although the convergence of the resulting algorithm is not proven, it has demonstrated empirical efficiency on both noiseless and noisy data for different types of applications on several numerical experiments.
In our setting, it leads to update $\lambda^{(h)}$ and $p^{(h)}$ at each $h$-th iteration as stated in Equation \eqref{eq:optim_Rp}, meaning that each $\TpV$ problem is stopped after only one iteration.

\section{The proposed AI-assisted incremental algorithm} \label{sec:proposed}

Despite the usage of inexact iterative solutions to lower the large number of iterations to converge to an optimal solution, the high computational cost of MB methods still remains a significant concern for mathematical approaches.
Their prolonged execution time makes them impractical for clinical applications, where real-time or near-real-time processing is crucial, and long processing times can hinder workflow efficiency. 

To address this challenge, neural networks (NNs) have emerged as a promising solution for {\it accelerating} medical image processing based on MB approaches. 
In fact, after an initial period of great hype driven by their outstanding performance on several imaging tasks, neural networks are now facing a phase of increased scrutiny. Concerns about hallucinations (i.e., where the model generates artifacts or misleading structures) and the inherent black-box nature of trained networks have raised doubts about the reliability of Deep Learning (DL) tools in clinical applications \cite{Arridge_Maass_Öktem_Schönlieb_2019, sidky2021docnn, morotti2021green,  gap_between_theory_and_practice, robustness_included, evangelista2023ambiguity}. \\
These challenges have made neural networks appear less appealing as standalone solutions, motivating the research shift towards {\it hybrid approaches} that integrate deep learning with traditional variational methods: instead of completely replacing model-based techniques, neural networks are increasingly being used to enhance and accelerate them. Nowadays, to set some examples, hybrid paradigms leverage the efficiency and accuracy of neural networks to embed learned priors in bi-level methods \cite{calatroni2017bilevel, bubbaCalatroniCatozziRautio_bilevel}, plug denoisers in multi-step algorithms \cite{venkatakrishnan2013plug, ahmad2020plug, cascarano2022plug, kamilov2023plug}, or compute optimal parameters for the MB solvers in unrolling or multi-step strategies \cite{monga2021algorithm, bubba2019learning, kofler2023learning}. 
It is well known that, unfortunately, this typically happens at the expense of some theoretical guarantees about the convergence of the variational methods, as the assumptions required on the networks are typically hard to verify in practice. 

In this Section, we delineate our proposal, which embeds pretrained neural networks within the inc\TpV framework, carefully designed to preserve the theoretical properties of the underlying optimization algorithm.
Different versions of the proposed approach will be presented, to fit different medical imaging scenarios.

\subsection{Incremental Deep Guess (incDG)}\label{ssec:incDG}

In the recent work \cite{LoliPiccolomini2025DeepGuess}, the Deep Guess (DG) strategy has been proposed for accelerating the execution of a MB solver that addresses non-convex minimization. 
It consists of using a deep neural network to generate a good approximation of the ground truth image (or the MB solution) and using it as the starting guess $\tilde{\x}$ for the execution of the MB solver. The authors remark that because of the high quality of the NN output, only a few iterations can be executed to achieve convergence solutions, that even report enhanced quality over the mere MB and the mere NN solutions. 

In this paper, therefore, we integrate the DG tool within each iteration of the incremental scheme. For this reason, we denote our proposal as {\it incDG}.\\
Precisely, once H is set, H convolutional neural networks $\Psi^{(h)}: \R^n \to   \R^n$ with $h = 0, 1, \dots, H-1$ are used incrementally, i.e., each $\Psi^{(h)}$ maps the solution $\bar{\x}^{(h)}$, computed as in \eqref{eq:outer_step} in the previous step, towards the target image, for any $h > 0$. The first network $\Psi^{(0)}$ operates on the starting guess $\tilde{\x} =: \bar{\x}^{(0)}$ image.
We denote the output of the network with $\hat{\x}^{(h+1)}$, such that:
\begin{equation}\label{eq:NN_h}
    \Psi^{(h)}: \bar{\x}^{(h)} \longmapsto \hat{\x}^{(h+1)}, \quad\forall\ h = 0, \dots, H-1.
\end{equation}
Subsequently, according to the DG strategy, we run a few iterations of the IR algorithm, to compute the numerical solution $\bar{\x^{(h+1)}}$ of the $h$-th non-convex problem with Algorithm \ref{alg:cp_tpv}. 
At last, recalling the incremental scheme of Algorithm \ref{alg:inc_tpv}, we update the model parameters and the algorithmic variables for the next iteration as described in Section \ref{ssec:incTpV}.

\begin{algorithm}[hbt!]
\caption{Incremental Deep Guess algorithm for $\TpV$ minimization (incDG)}\label{alg:inc_dg}
{\bf input:} $\K, \y, \lambda^{(0)}, \tilde{\x}, \alpha_p \in (0, 1), H \in \mathbb{N}, K \in \mathbb{N}^H$,  $\tau_x$, $\tau_\mathcal{F}$
\tcp{Initializations}
$p^{(0)} = 1, \bar{\x}^{(0)} = \tilde{\x}$; \\
$ f^{(0)} = \frac{1}{2} || \K\bar{\x}^{(0)} - \y ||_2^2 + \lambda^{(0)} || \> | \D\bar{\x}^{(0)}| \> ||_{p^{(0)}}^{p^{(0)}} $; \\
$h=0$; \\
\While{$h < H$}{  
    \tcp{Deep Guess step with warm starting}
    $\hat{\x}^{(h+1)} = \Psi^{(h)}(\bar{\x}^{(h)}) $\; 
    \tcp{Solve subproblem}
    Compute $\bar{\x}^{(h+1)} \in \arg\min_ {\x \in \Omega} \frac{1}{2}|| \K\x - \y ||_2^2 + \lambda^{(h)}\cdot ||\ |\D \x|\ ||_{p^{(h)}}^{p^{(h)}}$     by Algorithm \ref{alg:cp_tpv} with starting guess $\hat{\x}^{(h+1)}$, tollerances $\tau_x$ and $\tau_\mathcal{F}$, and ${\tt k}^{IR}=K_h$\;
    \tcp{Update parameters}
    $ f^{(h)} = \frac{1}{2} || \K\bar{\x}^{(h+1)} - \y ||_2^2 + \lambda^{(h)}\cdot || \ | \D\bar{\x}^{(h+1)} | \> ||_{p^{(h)}}^{p^{(h)}}$\;
    $p^{(h+1)} = p^{(h)} \cdot \alpha_p$\;
    {\bf if\ } $h=0${\bf :}   $\lambda^{(h+1)} = \lambda^{(0)} / 2$\;
    {\bf else\ } $\lambda^{(h+1)} = \lambda^{(h)} \cdot \frac{f^{(h)}}{f^{(h-1)}}$\;
} 
\Return $\bar{\x}^{(H)}$\;
\end{algorithm}

The entire incDG procedure is reported in Algorithm \ref{alg:inc_dg}.
Once the scheduler $K\in \N^H$ with the (limited) number of doable iterations is set, the proposed algorithm can be used with the same configuration setting of the inc$\TpV$ solver. 
Notably, the forward pass of a pre-trained neural network is highly efficient, requiring minimal computational time. As a result, the entire incDG procedure can be performed quickly.

\subsection{Training of the Deep Guess networks}\label{ssec:incDG_training}

All the networks are trained with a supervised approach, requiring a dataset with target images. 
In medical scenarios, however, the availability of ground truth images is not a given. 

Luckily, in applications such as denoising and deblurring, the goal is to learn estimations of parameters like noise statistics or blur magnitude. As a result, the training dataset does not always need to consist of images that are strictly coherent with the anatomy under examination. In fact, it can be designed to capture the relevant features of image degradation, allowing the network to focus on restoring image quality without learning specific image content.
For example, high-quality photographic pictures or synthetic images with geometric elements can be used as accurate ground truth samples \cite{cascarano2022plug}. 
In this case, we can thus assume to have a collection of $T$ high-quality images $\{ \x^{GT}_i \}_{i=1}^T$ and the corresponding $T$ corrupted images $\{ \y_i \}_{i=1}^T$. \\
For all our $H$ networks, we set the loss function as the Mean Squared Error (MSE) between the NN predictions and the corresponding $\x^{GT}_i $ samples. Consequently, the first image-to-image operator is defined  so that:
\begin{equation}\label{eq:Psi_0_toGT}
    \Psi^{(0)} = \arg\min_{\Psi} \sum_{i=1}^{T} || \Psi(\bar{\x}^{(0)}_i) - \x^{GT}_i ||_2^2,
\end{equation}
involving the prefixed starting points $\bar{\x}^{0}_i$, whereas the following $\Psi^{(h)}$ operators for $h=1, 2, \dots, H-1$, are given by:
\begin{equation}\label{eq:Psi_h_toGT}
    \Psi^{(h)} = \arg\min_{\Psi} \sum_{i=1}^{T} || \Psi(  \hat{\x}^{(h)}_i) - \x^{GT}_i ||_2^2.
\end{equation}
We remark that, in the training phase, each network $\Psi^{(h)}$ directly works on $\hat{\x}^{(h)}_i$ instead of on $\bar{\x}^{(h)}_i$, because we prefer not to fix the scheduler $K$ of iterations for training. \\


In applications addressing reconstruction from physical measurements, however, the networks must learn to distinguish the actual image content from the artifacts due to the underlying physics of data acquisition. 
In these cases, it is preferable to use target images with specific anatomical details that closely match the expected clinical data, because a mismatch between training and test images (such as using synthetic or overly simplified anatomical structures during training) can lead to poor generalization, making the network fail in real-world scenarios \cite{morotti2021green, mathematics_of_adversarial_attacks, solvability_of_inverse_problems_in_medical_imaging, yoon2023domain}.\\
In 2023, Evangelista et al. \cite{evangelista2023rising} introduced a ground-truth-free approach for training NNs without access to clean images. Their method proposes starting from available corrupted data and using the corresponding variational solutions, computed with high accuracy by a MB solver, as target samples. This framework has been formally analyzed as an inverse problem solver in \cite{evangelista2025or} and already applied to imaging tasks in \cite{evangelista2023ambiguity, morotti2023robust}. 
According to this idea, we run the inc$\TpV$ scheme saving all the incremental solutions $\bar{\x}_i^{(h)}$ for all $i=1; \dots, T$ and $h=1, \dots, H$, then we define our NN operators as:
\begin{equation}\label{eq:Psi_0_toIS}
    \Psi^{(0)} = \arg\min_{\Psi} \sum_{i=1}^{T} || \Psi(\bar{x}^{(0)}_i) - \bar{\x}^{(1)}_i ||_2^2,
\end{equation}
and:
\begin{equation}\label{eq:Psi_h_toIS}
    \Psi^{(h)} = \arg\min_{\Psi} \sum_{i=1}^{T} || \Psi(  \hat{\x}^{(h)}_i) - \bar{\x}^{(h+1)}_i ||_2^2, \qquad \forall h=1, 2, \dots, H-1.
\end{equation}

\subsection{Architecture of the NN}\label{ssec:UNet}

\begin{figure}
    \centering
    \includegraphics[width=0.75\linewidth]{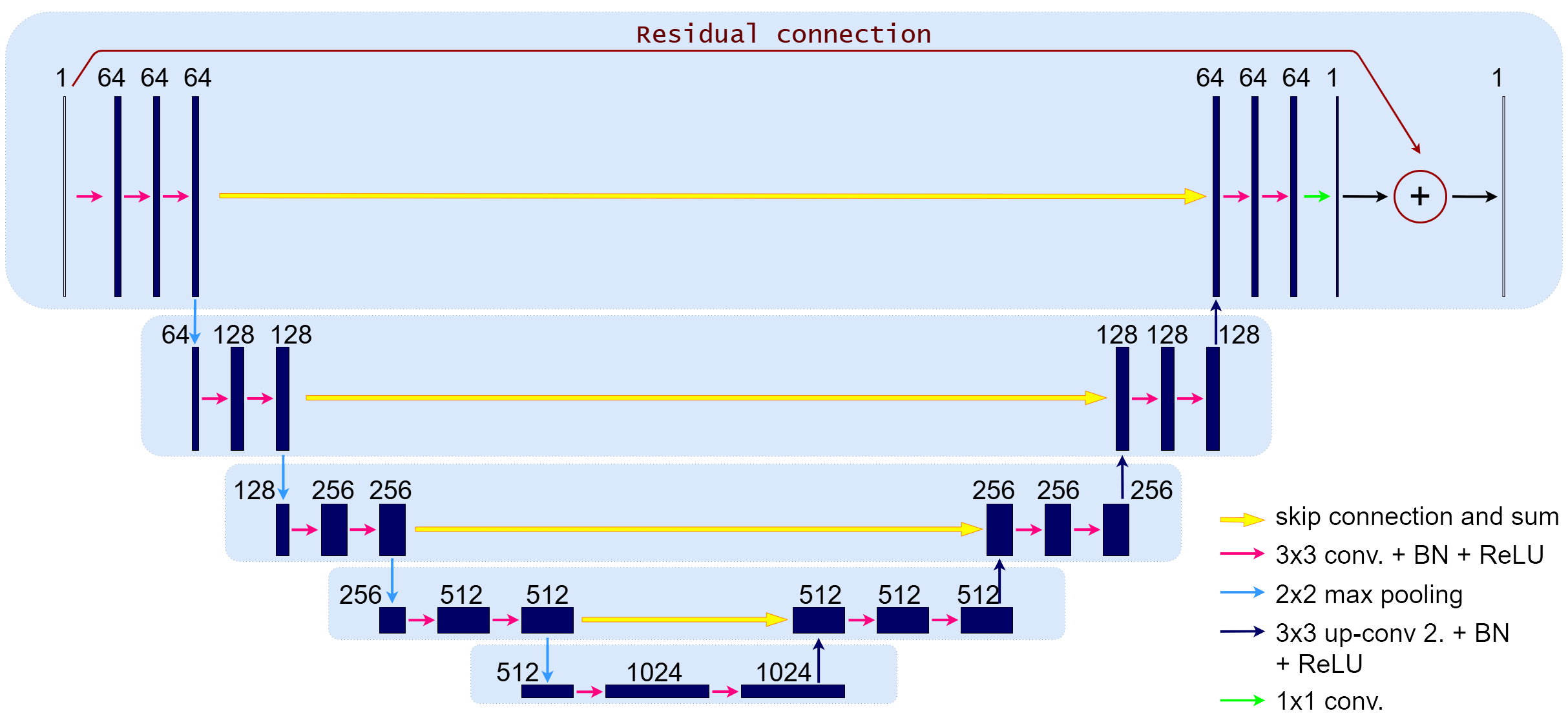}
    \caption{Scheme of the ResUNet architecture used to define the image-to-image operators $\Psi^{(h)}$ for the Deep Guess step. }
    \label{fig:resunet}
\end{figure}

In all our implementations, the neural network used in the Deep Guess step acts as an image-to-image operator in the image domain.
Its architecture is illustrated in Figure \ref{fig:resunet}. 
It is a fully convolutional network based on the  popular UNet architecture, popularized by \cite{ronneberger2015u}, with a symmetric encoder-decoder structure to input and output images of same dimensions. 
The encoder uses pooling operations to define resolution levels 
$ l = 0, \dots, L$, while the decoder mirrors this structure with upsampling layers.
Advances in image processing have shown that the UNet architecture is effective in handling global artifacts, because the pooling and unpooling mechanisms expand the receptive field of convolutional filters. 
Additionally, skip connections, implemented as element-wise additions to reduce parameters, preserve high-frequency information by linking corresponding encoder and decoder levels.
This makes UNet particularly suitable for image reconstruction from subsampled data, where streaking artifacts affect the entire image \cite{ye2018deep, you2019low, zioulis2022hybrid, zioulis2022hybrid}.\\
More precisely, we adopt the residual UNet (ResUNet) architecture, as implemented in \cite{wang2019admm}.
The residual connection proposed in \cite{han2016deep} is a shortcut path that directly links the input of the neural network layer to its output. Its presence improves gradient flow during training, mitigating the vanishing gradient, and leverages the fact that learning the residual map:
\begin{equation*} 
    \mathcal{M}_R : y 	\longmapsto y + \hat{x}
\end{equation*}
is often easier than directly learning the correction map: 
\begin{equation*} 
    \mathcal{M}_C : y 	\longmapsto \bar{x}.
\end{equation*}
Indeed, by focusing on predicting the difference between the input and the desired output, rather than reconstructing the entire image from scratch, the network can concentrate on modeling the artifacts, leading to more stable training and better generalization \cite{venkatesh2018deep, zhang2018road, han2018framing, jha2019resunet++, kumar2023brain}.

\section{Results} \label{sec:results}

In this Section, numerical experiments are reported to compare the proposed incDG framework to both variational MB and DL-based solvers, on real and synthetic medical images.
We consider two imaging inverse problems, i.e., image deblurring  and 2D image reconstruction from subsampled tomographic data, which represent two very different scenarios for medical imaging, as discussed in Section \ref{ssec:incDG_training}.


\subsection{Experimental setup} \label{ssec:results_setup}

Our experiments are based on numerical simulations and publicly available datasets.

In all cases, for each ground truth image $\x^{GT}_i$ in a dataset, we sampled a noise realization $\e_i \sim \mathcal{N}(\boldsymbol{0}, I)$, and we computed corrupted data as $ \y_i = \K \x^{GT}_i + \nu \frac{||\K \x^{GT}_i||}{||\e_i||} \e_i, $ where $\nu \geq 0 $ is the noise level.\\

For the image restoration task, we synthesize the blurring effect with the Gaussian degradation model, implemented with a $11 \times 11$ Gaussian kernel $\mathcal{G}$ given by: 
\begin{align}\label{eq:gaussian_kernel}
    \mathcal{G}_{i, j} = \begin{cases}
         e^{- \frac{1}{2} \frac{i^2 + j^2}{\sigma_G^2}} \quad & i, j \in \{-5, \dots, 5\}^2 \\
         0 & \text{otherwise}
    \end{cases}
\end{align}
with variance $\sigma_G = 1.3$.\\
For training the networks, we used 400 images from the COULE dataset\footnote{ \url{https://www.kaggle.com/datasets/loiboresearchgroup/coule-dataset}}. This dataset consists of grayscale images of size  256×256 pixels ($n=m=65536$) and contains overlapping ellipses having uniform intensities and different contrast to the background. 
For testing, 30 samples were used to evaluate the performance onto images coherent with the training ones. Additionally, 30 images from the `Brain CT' dataset\footnote{{\url{https://www.kaggle.com/datasets/vbookshelf/computed-tomography-ct-images}}} are considered, only for testing purposes. This dataset contains tomographic head reconstructions of patients with intracranial hemorrhages in the  brain tissue, and the associated mask images identifying the precise location of bleeding. In addition, these test images reflect the anatomical diversity of head CT, with images of the mandibular arch, teeth, and nasal cavity.  \\
All the experiments, reported in \ref{ssec:results_deblur}, have been achieved using added noise with $\nu=0.02$. \\

For the CT reconstruction problem, we used the Mayo Clinic dataset, constituted of images of authentic human abdomens \cite{mccollough2016tu}. The images have been downsized to $256 \times 256$ pixels ($n=65536$) and scaled to the $[0, 1]$ range. We used 1000 images for training and 150 for testing.
The linear operator $\K$ has been computed with Astra Toolbox \cite{van2015astra,van2016fast} functions, simulating a fan beam sparse geometry with $60$ projections uniformly distributed within the angular range $[0, 180)$ degrees, with a detector resolution of $500$ pixels (hence $m=30000$). \\ 
In CT reconstruction experiments, we added noise with $\nu=0.005$. The corresponding solutions are reported and discussed in \ref{ssec:results_ct}.\\

To execute the inc$\TpV$ solver for deblurring, we always set the initial guess $\x^{(0)}$ as the corrupted image $\y$, 
$H=4$ with scheduler $K = [100, 100, 50, 10]$ for computing numerical convergence solutions $\bar{\x}^{(h)}$ (hence $\bar{K}=270$), $\alpha_p = 0.5$ to update the sparsity parameter according to \eqref{eq:UpdateP},  and $\lambda^{(0)}=0.5$ to initialize the sequence of regularization parameters as in \eqref{eq:UpdateLambda}. \\
When running the proposed inc$\TpV$ for image reconstruction,  the starting guess $\x^{(0)}$ is always set as the Filtered Back Projection (FBP) solution \cite{beylkin1987discrete}, which can be computed from the $\y$ sinogram data in a very short time and still represents a benchmark for CT imaging \cite{pan2009commercial}. 
Additionally, we use $H=6$ with scheduler $K =[200, 500, 500, 500, 700, 700]$ (hence $\bar{K}=3100$), $\alpha_p = 0.7$ and $\lambda^{(0)}=0.01$.\\
The CP solver always make use of $\tau_x = 10^{-7}$, $\tau_{\mathcal F} = 10^{-7}$, {\tt k}$^{CP} = 5$, and $\xi = 2\cdot10^{-3}$. \\
All these values have been set to achieve good performance on the training samples, on average. \\

To implement the proposed incDG algorithm, we trained the networks as presented in Section \ref{ssec:incDG_training}. 
Specifically, we use only the neural networks defined in Equations \eqref{eq:Psi_0_toGT} and \eqref{eq:Psi_h_toGT} in case of image deblurring, whereas for CT reconstructions we also use networks as in Equations \eqref{eq:Psi_0_toIS} and \eqref{eq:Psi_h_toIS}.
The training phase of each network used the Adam optimizer with a learning rate of $0.001$ and performed 100 epochs.
The parameters used in inc$\TpV$ remain unchanged, except for the scheduler $K$. We set here  $K =[5, 5, 5, 5]$ (hence $\bar{K}=20$) for image deblurring, whereas $K =[5, 5, 5, 5,5,5]$ ($\bar{K}=30$) is used for CT reconstruction. \\

In the following, we compare the incDG and the inc$\TpV$ solutions to those obtained using the $\TpV$ model, which serves as our primary state-of-the-art benchmark.
We thus consider the MB convergence images computed via the IR Algorithm \ref{alg:cp_tpv} addressing the $\TpV$ model with fixed $\lambda$ parameter, heuristically set on the training samples. We also consider a variant that uses the decreasing sequence of regularization parameters defined with the updating rule in Equation \ref{eq:UpdateLambda}.
In both cases, we set $p$ equal to the last parameter $p^{(H-1)}$ used in the inc$\TpV$ and incDG solvers, and {\tt k}$^{IR} = 270$ matching $\bar{K}$ for image deblurring, or {\tt k}$^{IR} =1000$ for CT application.\\
We also introduce incNN, a variant of incDG where no MB iterations are performed during the incremental steps. This corresponds to setting a null scheduler, so that the trained networks are simply concatenated, and the final output lacks a formal mathematical characterization.\\

Besides visual inspection and zooms-in on regions of particular medical interest, the comparison of results is based on quality assessment metrics. 
We use both the Relative Error (RE) and the Structural Similarity Index (SSIM) \cite{wang2003multiscale}, as they capture different aspects of reconstruction quality.
RE measures the pixel-wise discrepancy between the computed reconstruction and the ground truth, providing a quantitative evaluation of overall accuracy in solving the imaging inverse problems. However, it does not account for perceptual quality or structural integrity. 
SSIM, on the other hand, evaluates image similarity in terms of luminance, contrast, and structural information, making it more sensitive to the preservation of anatomical details. 
The lower the RE and the higher the SSIM, the better the image quality, ensuring a balance between numerical accuracy and visual fidelity.
For the CT reconstruction application, we also compare the solvers in terms of computational times, since fast reconstruction is often crucial in clinical practice, where near real-time imaging directly impacts diagnostic workflows \cite{cavicchioli2020gpu}.

\subsection{Image deblurring and denoising} \label{ssec:results_deblur}

We now tackle the image enhancement problem of deblurring and denoising images.
Firstly, we evaluate the incremental strategies common to the inc$\TpV$, incDG, and incNN approaches, then we focus on the comparison of results on the entire test set with COULE samples, in Section \ref{ssec:results_deblur_increm}. 
Then, in Section \ref{ssec:results_deblur_brain}, we assess the generalization capability of incDG on unseen images from the Brain CT dataset.

\subsubsection{Incremental strategies and their generalizability on the testing dataset}\label{ssec:results_deblur_increm}
\begin{figure}
    \centering
    \renewcommand{\arraystretch}{0.5} 
    \begin{tabular}{c ccc}
    $\x^{GT}$ & $\y$ & $\TpV$ (fixed $\lambda$) & $\TpV$ \\ \rule{0pt}{4mm}
        - & 
        RE = 0.2673 & 
        RE = 0.2127 & 
        RE = 0.0920  \\ 
        - &
        SSIM = 0.6430 & 
        SSIM = 0.7796 & 
        SSIM = 0.9376 \\ 
         \begin{tikzpicture}
            \node [anchor=south west, inner sep=0] (image) at (0,0) 
            {\includegraphics[width=0.22\linewidth]{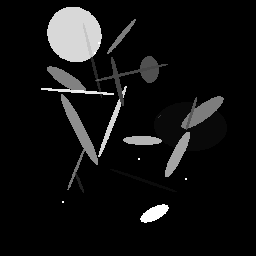}};
            \begin{scope}[x={(image.south east)}, y={(image.north west)}]
                \draw[red, thick] (0.41, 0.35) rectangle (0.91, 0.65);
            \end{scope}
        \end{tikzpicture} &
        \includegraphics[width=0.22\linewidth]{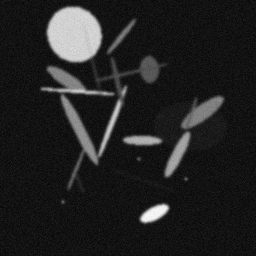} &
        \includegraphics[width=0.22\linewidth]{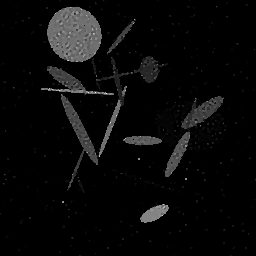} &
        \includegraphics[width=0.22\linewidth]{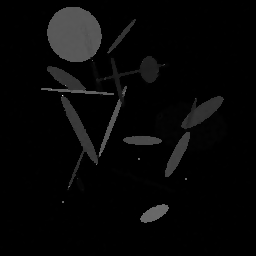} \\ 
        \begin{tikzpicture}
            \node [anchor=south west, inner sep=0] (image) at (0,0) 
            {\includegraphics[clip, trim = {25mm 22mm 5mm 23mm},   width=0.22\linewidth]{coule__gt_11.png}};
            \begin{scope}[x={(image.south east)}, y={(image.north west)}]
                \draw[yellow, -to](0.38, 0.98) -- (0.47, 0.79);  
                \draw[yellow, to-](0.86, 0.28) -- (0.95, 0.15);
                \draw[yellow, to-](0.67, 0.20) -- (0.67, 0.05);
                \draw[blue, to-](0.70, 0.70) -- (0.85, 0.65); 
                
            \end{scope}
        \end{tikzpicture}  
         & 
        \includegraphics[clip, trim = {25mm 22mm 5mm 23mm}, width=0.22\linewidth]{coule__y_11.png}  & 
        \includegraphics[clip, trim = {25mm 22mm 5mm 23mm}, width=0.22\linewidth]{coule__tpv_11.png} &
        \includegraphics[clip, trim = {25mm 22mm 5mm 23mm}, width=0.22\linewidth]{coule__tpv_alphaP1_11.png}\\  \rule{0pt}{7mm}
    \end{tabular} 
    \begin{tabular}{c cc}
        inc$\TpV$ & incNN & incDG \\ \rule{0pt}{4mm}
        RE = 0.0951     & RE = 0.0729 & RE = 0.0531 \\ 
        SSIM = 0.9324  &  SSIM = 0.9816 & SSIM = 0.9906  \\
        \includegraphics[width=0.22\linewidth]{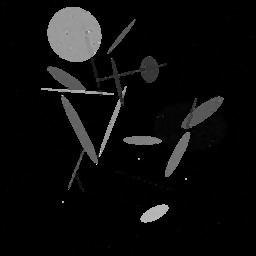} 
        & \includegraphics[width=0.22\linewidth]{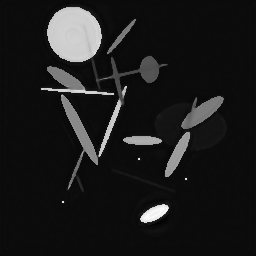} 
        & \includegraphics[width=0.22\linewidth]{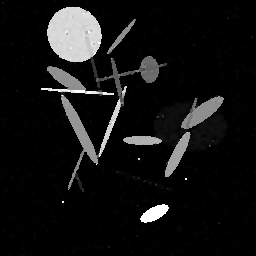} \\
        \includegraphics[clip, trim = {25mm 22mm 5mm 23mm}, width=0.22\linewidth]{coule__incTpV_11.png}
        & \includegraphics[clip, trim = {25mm 22mm 5mm 23mm}, width=0.22\linewidth]{coule__incNN_11.png}
        & \includegraphics[clip, trim = {25mm 22mm 5mm 23mm}, width=0.22\linewidth]{coule__incDG_11.png}
    \end{tabular}
    \caption{Experiments performed on a COULE test sample for image deblurring and denoising. On the top, from left to right: the ground truth image with a red rectangle depicting the zoomed area, containing a very low contrast ellipse marked by yellow arrows and a subtle line indicated by a blue arrow; the simulated $\y$ image, affected by blur and noise; the solutions of the $\TpV$-regularized model with fixed or variable regularization weights. On the bottom: the solutions obtained using inc$\TpV$, incNN and incDG schemes, with the corresponding zoomed regions of interest.} \label{fig:coule_11_metodi}
\end{figure}

In the upper part of Figure \ref{fig:coule_11_metodi}, one ground truth image of the COULE test set is visualized, together with its corrupted version $y$ and the TpV solutions (computed with $p=0.125$). 
The corresponding final solutions achieved by inc$\TpV$, incNn and incDG methods are shown in the lower part of Figure \ref{fig:coule_11_metodi}.
The RE and SSIM values are reported over each image, whereas a region of interest is zoomed, depicting very low-contrast elements that are pointed by arrows.
In Table \ref{tab:Coule_h}, mean and standard deviation values of RE and SSIM metrics computed on the entire test set are reported for each incremental step of inc$\TpV$, incNN and incDG, and for the corrupted input images $\y$.

As visible, the $\TpV$ solution computed with fixed $\lambda$ is well deblurred but some noisy components have been magnified by the regularized approach. On the contrary, using a decreasing sequence of regularization parameters has boosted the denoising effect remarkably. Both the $\TpV$ solutions, however, appear poorly contrasted and quite dark. 
All the images by the incremental approaches are lighter, well-denoised and restored, as also proven by their metrics. The final solutions by incNN and incDG hit impressive SSIM scores.
Furthermore, incNN perfectly discerns the large low-contrast ellipse (pointed in yellow on the ground truth image) from the background, whereas inc$\TpV$ and incDG better restore the boundaries of the vertical low-contrast line (pointed on the ground truth image in blue).\\
Analyzing the metrics of the early solutions across incremental iterations reveals some noteworthy patterns. Notably, incNN achieves a very low RE on average after just one iteration, reaching its optimal performance at $h=1$ before degrading thereafter. This suggests that incNN does not effectively exploit the incremental scheme, as further iterations fail to refine the results.
In contrast, the inc$\TpV$ scheme benefits from additional incremental steps. To illustrate this, Table \ref{tab:Coule_h} includes an extra column for $h=4$, showing that the RE continues to decrease as iterations progress. However, the improvement is marginal and does not compensate for the additional computational cost; thus, all subsequent results are reported for $h=3$.
Finally, the proposed incDG algorithm strikes an effective balance between inc$\TpV$ and incNN: it quickly attains strong performance like incNN while also exploiting the incremental process of inc$\TpV$. 

\begin{table}[]
    \centering
    \begin{tabular}{ll ccccc}
       & & h=0 & h=1 & h=2 & h=3  & h=4\\
       \toprule
    \multirow{ 3}{*}{ \shortstack[l]{RE} } 
      & incTpV & 0.135$\pm$0.021 &  0.118$\pm$0.017 & 0.090$\pm$0.016 & 0.084$\pm$0.016   & 0.083$\pm$0.014\\ 
      & incNN &  0.055$\pm$0.014 & 0.052$\pm$0.014 & 0.055$\pm$0.014 & 0.056$\pm$0.014  \\
      & incDG & 0.060$\pm$0.012 & 0.052$\pm$0.012 & 0.049$\pm$0.012 & 0.048$\pm$0.012    \\ 
      & $\y$ & 0.246$\pm$0.024\\ 
       \midrule
    \multirow{ 3}{*}{ \shortstack[l]{SSIM} }    
      & incTpV & 0.904$\pm$0.006 & 0.913$\pm$0.005 &  0.929$\pm$0.003 & 0.933$\pm$0.003  & 0.933$\pm$0.004 \\
      & incNN & 0.888$\pm$0.026 & 0.963$\pm$0.036 & 0.881$\pm$0.108 & 0.850$\pm$0.123   \\
      & incDG & 0.976$\pm$0.005 & 0.985$\pm$0.003 & 0.986$\pm$0.003 & 0.986$\pm$0.002   \\
      & $\y$ & 0.650$\pm$0.040 \\
     \bottomrule
    \end{tabular}
    \caption{Experiments performed on a COULE test sample for image deblurring and denoising. Mean and Standard Deviation values of the assessment metrics computed for the incremental solutions at varying $h$ for the three incremental schemes considered in this study.}
    \label{tab:Coule_h}
\end{table}

\begin{figure}
    \centering
    \renewcommand{\arraystretch}{0.5} 
    \begin{tabular}{@{}cccc@{}}
    $\x^{GT}$ & incTpV & incNN & incDG \\ \rule{0pt}{5mm}
        - & 
        RE = 0.0852 & 
        RE = 0.0670 & 
        RE = 0.0398 \\ 
        - & 
        SSIM = 0.9261 & 
        SSIM = 0.4403 & 
        SSIM = 0.9777 \\ 
        %
        \begin{tikzpicture}
            \node [anchor=south west, inner sep=0] (image) at (0,0) 
            {\includegraphics[width=0.22\linewidth]{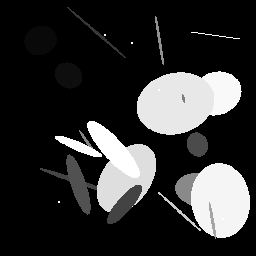}};
            \begin{scope}[x={(image.south east)}, y={(image.north west)}]
                \draw[red, thick] (0.08, 0.65) rectangle (0.61, 0.95);
                \draw[red, thick] (0.47, 0.09) rectangle (1, 0.39);
            \end{scope}
        \end{tikzpicture} & 
        \includegraphics[width=0.22\textwidth]{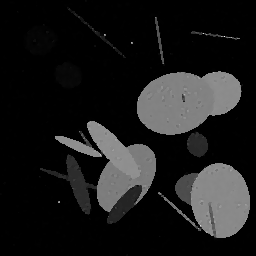} & 
        \includegraphics[width=0.22\textwidth]{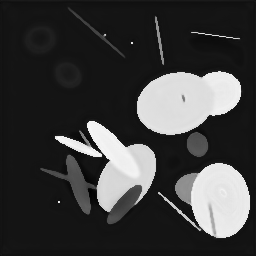} & 
        \includegraphics[width=0.22\textwidth]{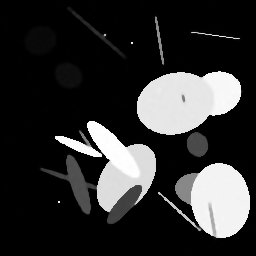} \\ 
        \includegraphics[clip, trim = {5mm 40mm 25mm 5mm}, width=0.22\textwidth]{coule__gt_4.png} &  
        \includegraphics[clip, trim = {5mm 40mm 25mm 5mm}, width=0.22\textwidth]{coule__incTpV_4.png} & 
        \includegraphics[clip, trim = {5mm 40mm 25mm 5mm}, width=0.22\textwidth]{coule__incNN_4.png} & 
        \includegraphics[clip, trim = {5mm 40mm 25mm 5mm}, width=0.22\textwidth]{coule__incDG_4.png} \\ 
        \includegraphics[clip, trim = {30mm 5mm 0mm 40mm}, width=0.22\textwidth]{coule__gt_4.png} &  
        \includegraphics[clip, trim = {30mm 5mm 0mm 40mm}, width=0.22\textwidth]{coule__incTpV_4.png} & 
        \includegraphics[clip, trim = {30mm 5mm 0mm 40mm}, width=0.22\textwidth]{coule__incNN_4.png} & 
        \includegraphics[clip, trim = {30mm 5mm 0mm 40mm}, width=0.22\textwidth]{coule__incDG_4.png} \\  
        \multicolumn{2}{c}{\includegraphics[width=0.45\textwidth]{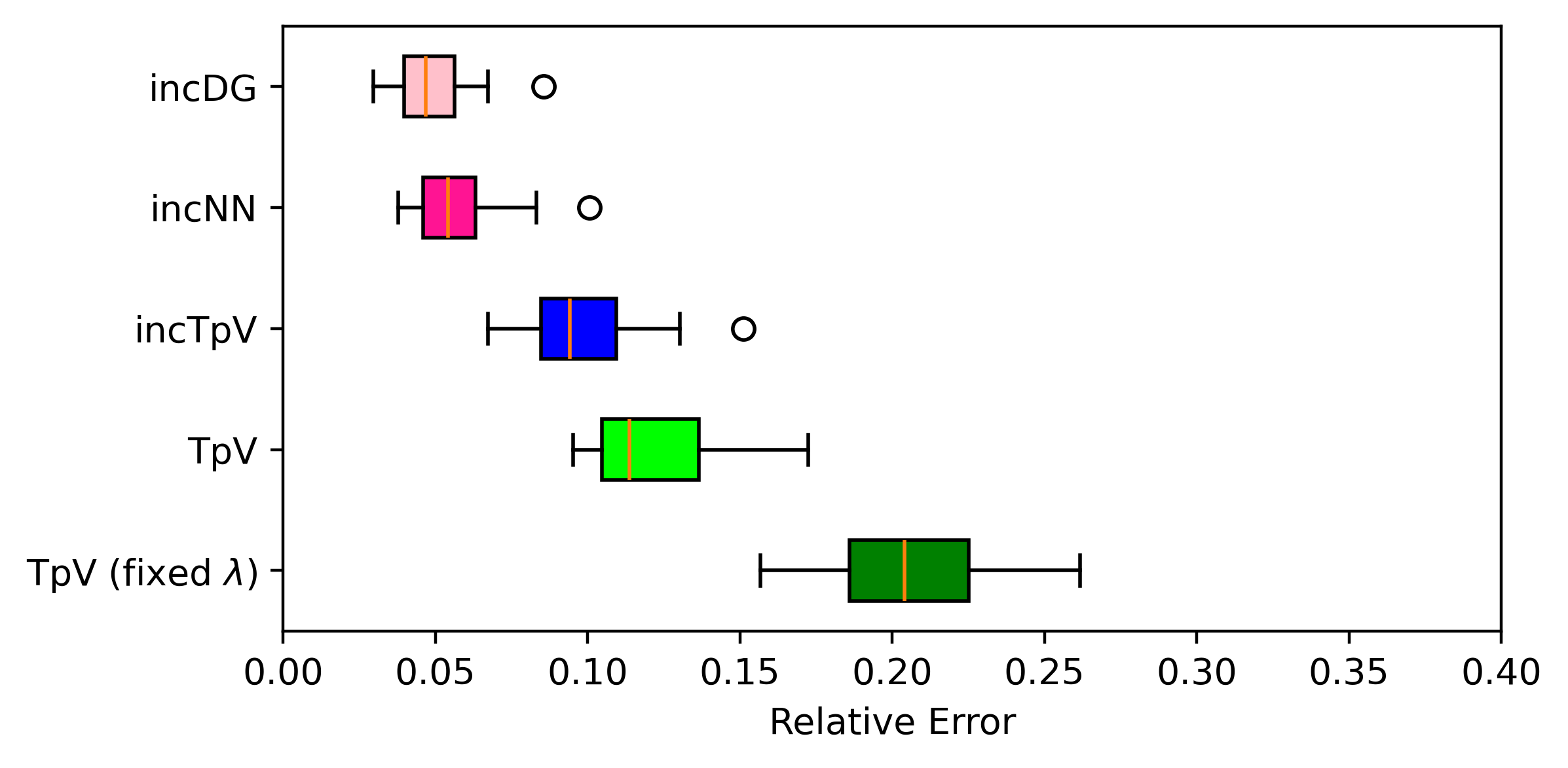} } &
        \multicolumn{2}{c}{\includegraphics[width=0.45\textwidth]{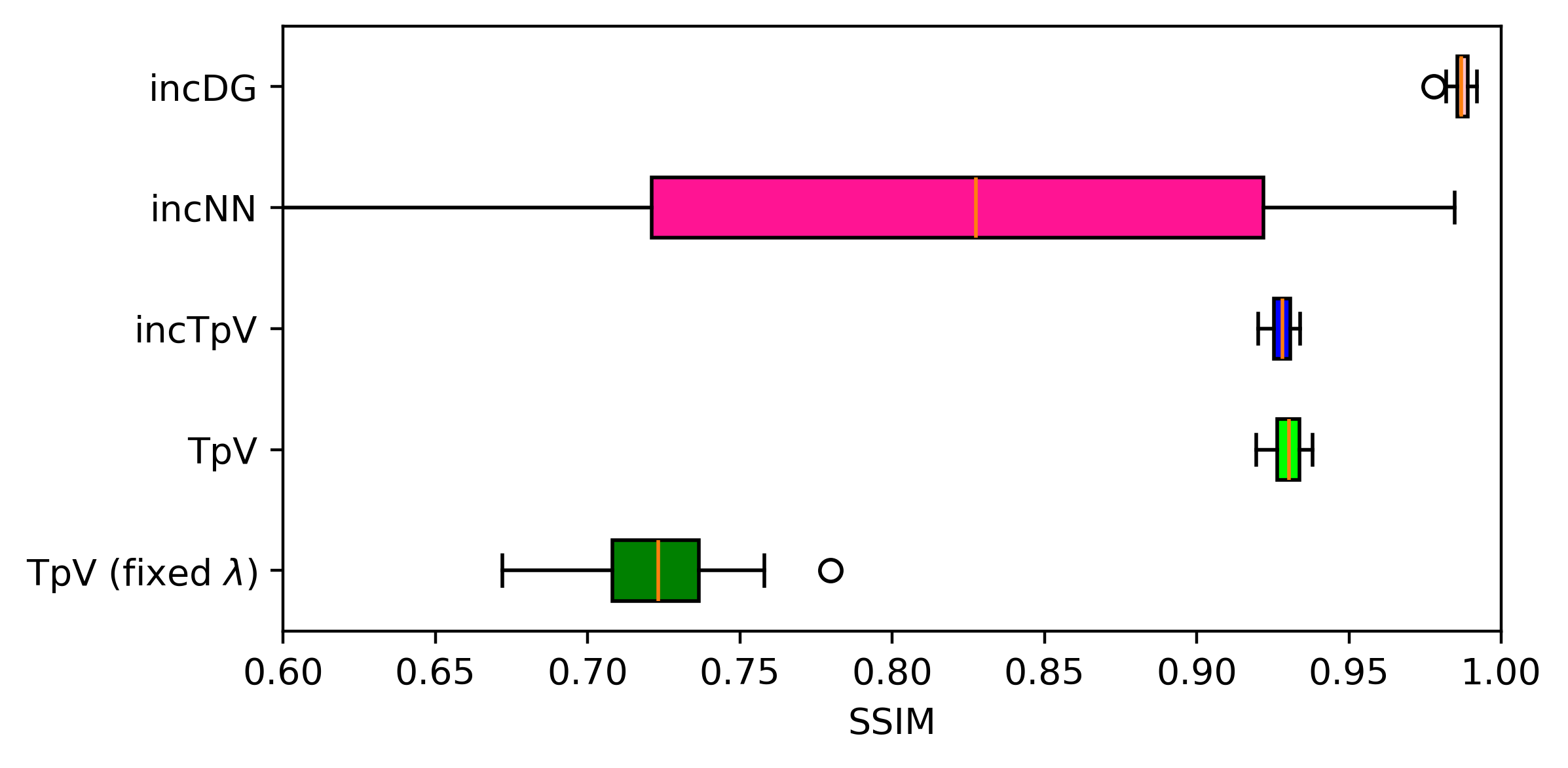} }\\
    \end{tabular}
    \caption{Experiments performed on COULE test samples for image deblurring and denoising. At the top: one ground truth image and the solutions obtained using inc$\TpV$, incNN and incDG schemes, with the corresponding zoomed regions of interest. At the bottom: boxplot comparisons of the RE metrics (left) and SSIM metrics (right), computed over the entire test set. }
    \label{fig:coule_19_4}
\end{figure}

Now, we further analyze the property of generalizability of the considered algorithms, which are used on the entire test set without changes in the parameter configuration. 
As previously mentioned, in clinical scenarios, fine-tuning of algorithmic parameters for each image is not feasible, hence the robustness of the input setting and the automatic updating rules must be carefully analyzed.
To this purpose, Figure \ref{fig:coule_19_4} offers useful insights.\\
At the top of Figure \ref{fig:coule_19_4}, we visually inspect the restoration of an additional test image processed by the inc$\TpV$, incNN, and incDG algorithms. 
While inc$\TpV$ and incDG achieve again excellent restorations, 
the incNN approach produces a solution with noticeable artifacts and poor similarity to the ground truth image. As highlighted in the zoomed-in regions, the two faint, dark ellipses in the first crop appear too light and lack uniform intensity, as they are overly smooth, with dark central areas. Similarly, in the second crop, the large bright ellipse reveals concentric artifacts, while other light structures display smoothed boundaries. Interestingly, the RE metric does not capture these inaccuracies like the SSIM value does.

Figure \ref{fig:coule_19_4} also presents a statistical analysis of the RE and SSIM metrics computed over the entire test set using boxplots, providing deeper insights for the performance of the solvers.
First, we observe that the incremental MB model (blue boxplots) outperforms the standard $\TpV$ algorithm significantly (dark green boxplots) as well as the algorithmic variant with variable regularization weights (light-green boxplots), confirming the effectiveness of the baseline inc$\TpV$ approach. 
Second, disregarding computational time differences between inc$\TpV$ and incDG, the advantage introduced by the deep guess strategy is evident. In fact, incDG consistently achieves superior performance, significantly surpassing inc$\TpV$ in both metrics.
Third, the unreliable behaviour of incNN is clearly reflected in the SSIM distribution. While the RE values remain relatively low, the SSIM scores exhibit extreme variability, ranging from values below 0.60 to over 0.98. As noted in the Section \ref{sec:proposed}, such fluctuations in NN-based approaches are well-documented in the literature and are commonly referred to as {\it instability}. 
Finally, we observe that, despite relying on the same neural networks as incNN, the proposed incDG approach is not affected by instability. The inclusion of a few MB iterations stabilizes the framework, as evidenced by the narrow spread of the (light-pink) boxplots, indicating minimal variability in the metric values.

\subsubsection{Generalizability on unseen medical images}\label{ssec:results_deblur_brain}

To further assess the generalization capability of the algorithms, we now evaluate their performance on the Brain CT dataset. Notably, the neural networks have not been re-trained but are the same as those used in the previous experiments.

\begin{figure}
    \renewcommand{\arraystretch}{0.5} 
    \begin{tabular}{c@{\hskip 0.5mm} c@{\hskip 0.5mm} c@{\hskip 0.5mm} c@{\hskip 0.5mm} c@{\hskip 0.5mm} c}
    $\x^{GT}$ & $\x^{GT}$ & mask & inc$\TpV$ & incNN & incDG \\ 
    \begin{tikzpicture}
            \node [anchor=south west, inner sep=0] (image) at (0,0) 
            {\includegraphics[width=0.15\textwidth]{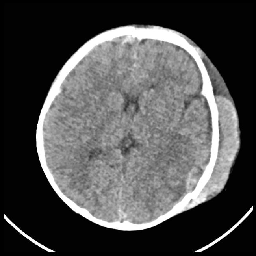} };
            \begin{scope}[x={(image.south east)}, y={(image.north west)}]
                \draw[red, thick] (0.65, 0.18) rectangle (0.90, 0.43);
            \end{scope}
        \end{tikzpicture} 
        & 
        \includegraphics[clip, trim = {42mm 10mm 5mm 37mm},  width=0.15\textwidth] {brain__gt_6.png} &
        \includegraphics[clip, trim = {42mm 10mm 5mm 37mm},  width=0.15\textwidth]{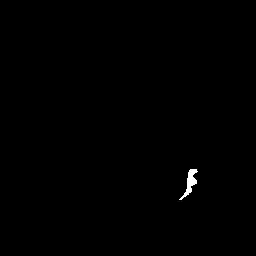}&
        \includegraphics[clip, trim = {42mm 10mm 5mm 37mm}, width=0.15\textwidth]{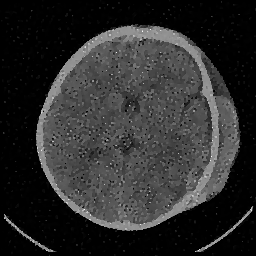} & 
        \includegraphics[clip, trim = {42mm 10mm 5mm 37mm}, width=0.15\textwidth]{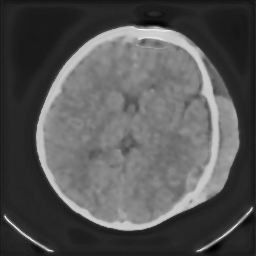} & 
        \includegraphics[clip, trim = {42mm 10mm 5mm 37mm}, width=0.15\textwidth]{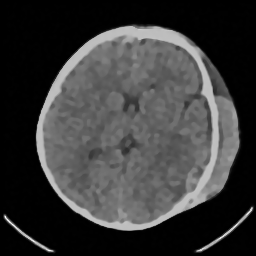} \\ \rule{0pt}{5mm} 
    \end{tabular}
    \centering
       \includegraphics[width=0.45\textwidth]{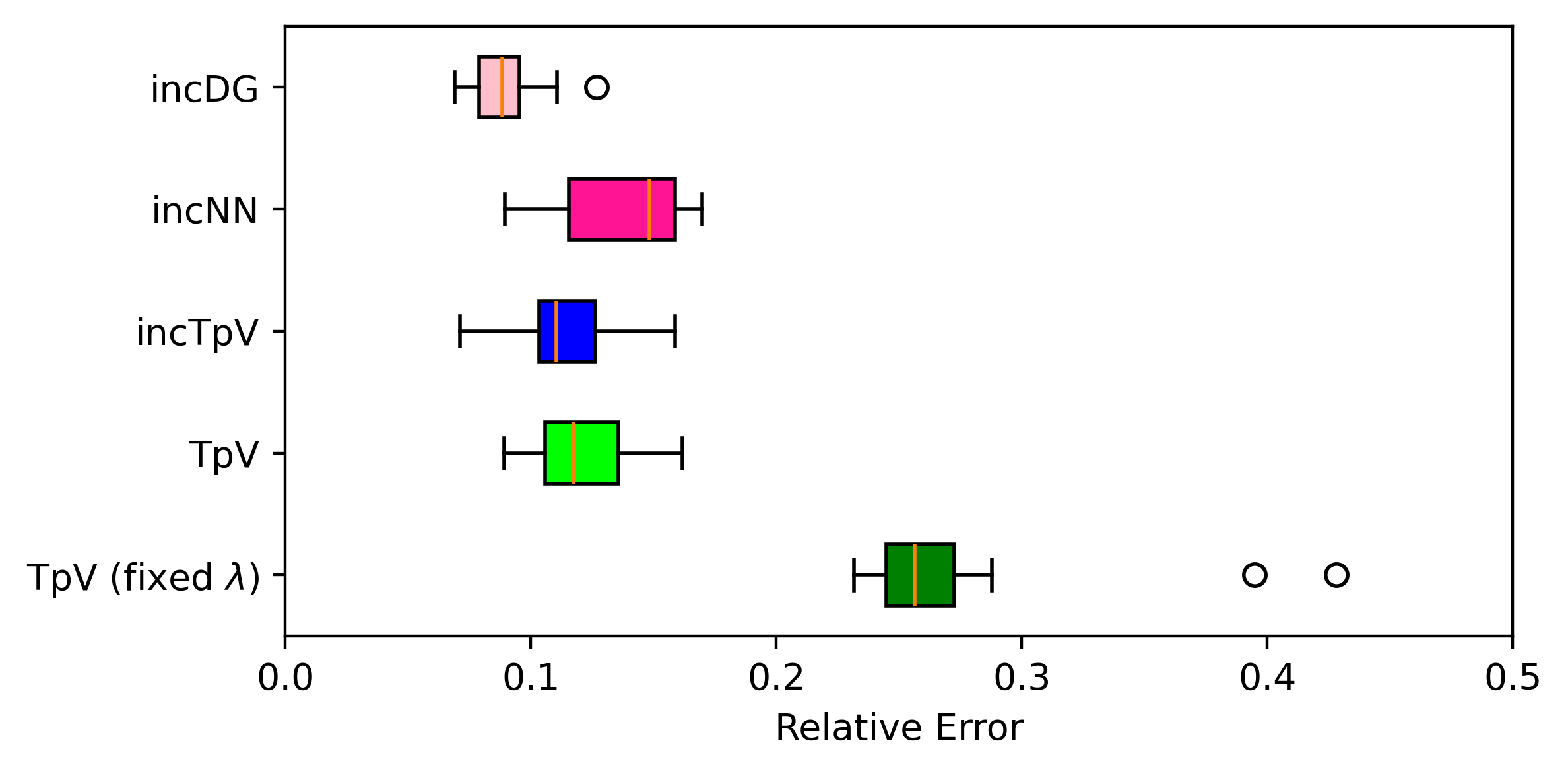} \hspace{-1mm}
 \includegraphics[width=0.45\textwidth]{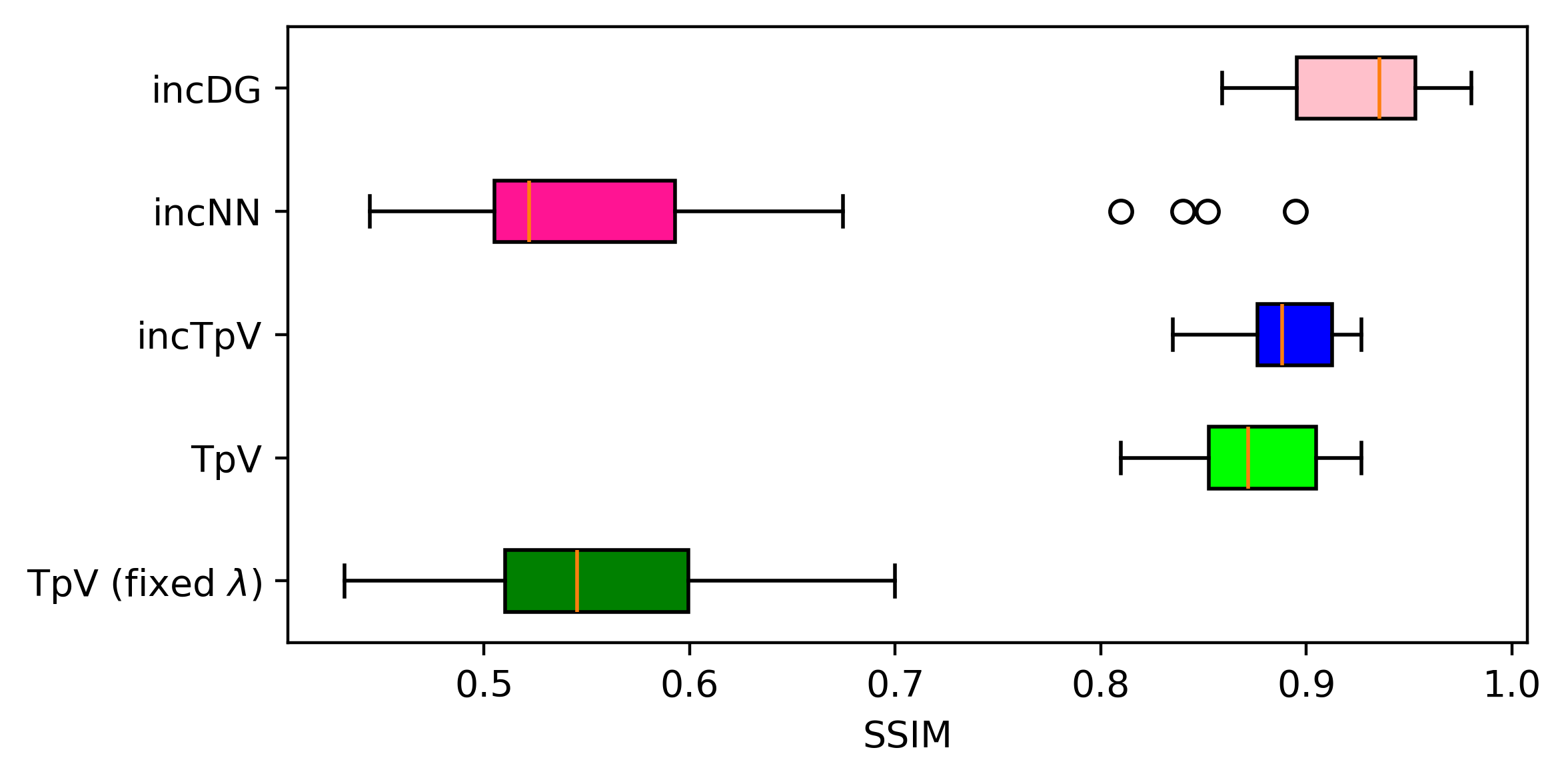}
    \caption{Experiments performed on Brain CT test samples for image deblurring and denoising. At the top, from left to right: one ground truth image with an intracranial hemorrhage; the zoomed-in area depicted by the red square on the previous image; the zoomed-in hemorrhage mask provided by the dataset; the zoomed-in areas on the solutions obtained using the inc$\TpV$, incNN and incDG schemes. At the bottom: boxplot comparisons of the RE metrics (left) and SSIM metrics (right), computed over the entire test set.}
    \label{fig:brain}
\end{figure}

Figure \ref{fig:brain} presents a test image containing an intracranial hemorrhage, within the red square. The lesion is more clearly visible in the zoomed-in crop (second image) and in the corresponding hemorrhage mask (third image), which is provided in the dataset as a reference. For clarity, Figure \ref{fig:brain} displays only the solutions obtained by the incremental schemes, while the RE and SSIM boxplots are also relative to the other considered solvers.

Examining the reconstructions, we first observe that inc$\TpV$ appears to have over-regularized the solution, suggesting that a different parameter configuration might have been beneficial. This further highlights the challenges associated with tuning MB methods in clinical scenarios.
Conversely, incNN produces a noticeably blurred reconstruction in this test, while incDG yields a more defined representation of the hemorrhage boundaries, which are not entirely evident even in the ground truth image. 

Focusing on the distributions of the metrics over 30 test samples, we stress that our incDG always outperforms the competitors, yielding the best median values, small fluctuations, and good metrics on the entire test set.
Additionally, we deduce that both the NN-based approaches do not manifest overfitting on the COULE images, as the charts are comparable to those of Figure \ref{fig:coule_19_4}.

\subsection{Tomographic reconstruction from subsampled data} \label{ssec:results_ct}

\begin{figure}
    \centering
    \begin{tikzpicture}
            \node [anchor=south west, inner sep=0] (image) at (0,0) 
            {\includegraphics[width=0.19\textwidth]{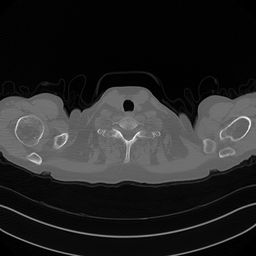} };
            \begin{scope}[x={(image.south east)}, y={(image.north west)}]
                \draw[red, thick] (0.32, 0.25) rectangle (0.72, 0.65);
            \end{scope}
        \end{tikzpicture}   
    \begin{tikzpicture}
            \node [anchor=south west, inner sep=0] (image) at (0,0) 
            {\includegraphics[width=0.19\textwidth]{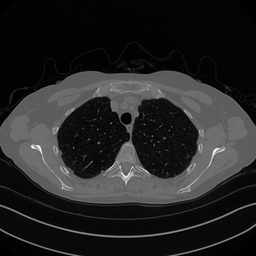} };
            \begin{scope}[x={(image.south east)}, y={(image.north west)}]
                \draw[red, thick] (0.29, 0.38) rectangle (0.69, 0.78);
            \end{scope}
        \end{tikzpicture}     
    \begin{tikzpicture}
            \node [anchor=south west, inner sep=0] (image) at (0,0) 
            {\includegraphics[width=0.19\textwidth]{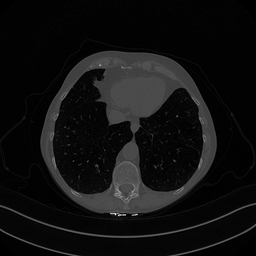} };
            \begin{scope}[x={(image.south east)}, y={(image.north west)}]
                \draw[red, thick] (0.50, 0.45) rectangle (0.90, 0.85);
            \end{scope}
        \end{tikzpicture}  
    \begin{tikzpicture}
            \node [anchor=south west, inner sep=0] (image) at (0,0) 
            {\includegraphics[width=0.19\textwidth]{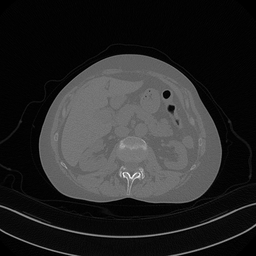} };
            \begin{scope}[x={(image.south east)}, y={(image.north west)}]
                \draw[red, thick] (0.42, 0.42) rectangle (0.82, 0.82);
            \end{scope}
        \end{tikzpicture}  
    \\ 
    \includegraphics[clip, trim = {23mm 17mm 17mm 23mm}, width=0.19\textwidth]{gt_7} 
    \includegraphics[clip, trim = {20mm 25mm 20mm 15mm}, width=0.19\textwidth]{gt_2.png} 
    \includegraphics[clip, trim = {35mm 30mm 5mm 10mm}, width=0.19\textwidth]{gt_237.png}   
    \includegraphics[clip, trim = {28mm 28mm 12mm 12mm}, width=0.19\textwidth]{gt_326.png}   
    \caption{Experiments performed on Mayo Clinic test samples for CT image reconstruction from subsampled data. Four ground truth images showing very different human anatomies (top), with zoomed regions of interest (bottom).  }
    \label{fig:ct_mayoGT}
\end{figure}

We conclude the numerical assessment of our proposal tackling tomographic image reconstructions.
In this case, we use real images provided by the Mayo Clinic hospital both for training and for testing. 
Some test images, corresponding to different slices of a patient's thorax, chest, and abdomen, are visible in Figure \ref{fig:ct_mayoGT}. Here, some zoomed regions highlight vertebrae, lungs and major blood vessels (aorta and pulmonary artery), heart and rib cage sections, and abdomen organs.\\

To further evaluate the versatility and practical impact of our method, we address a task where computational efficiency is particularly critical and minimizing execution time is essential.
Thus, before examining the reconstruction quality, we first focus on the mean running times, reported in Table \ref{tab:Mayo_times}, to reconstruct one single image by the tested solvers. 
The experiments were conducted on a Windows 10 Pro for Workstations system equipped with an Intel Xeon W-2223 CPU (3.60 GHz), 32 GB of RAM, and an NVIDIA RTX A4000 GPU (16 GB VRAM, 48 multiprocessors). The code was run using PyTorch 2.2.2 with CUDA 11.8 and cuDNN 8.7. 
In the table, the maximum number of performed CP iterations is reported for each solver. A clear advantage emerges for incDG, which achieves results in just 4.5 seconds, compared to 53.271 seconds required by the purely variational inc$\TpV$ approach.  Notably, since only 30 iterations are performed, a substantial portion of the total time (around 3.5 seconds, matching the incNN average) is spent on initializing the algorithms, loading data and network weights, and executing the network’s forward pass. \\

\begin{table}[]
    \centering
    \begin{tabular}{l ccccc}
       & FBP & $\TpV$ & inc$\TpV$ & incNN & incDG  \\
       \toprule
       Mean (secs) & 0.011 & 16.976 & 53.271 & 3.476 & 4.517 \\ 
       Iters. & - & 1000 & 3100 & - & 30 \\
     \bottomrule
    \end{tabular}
    \caption{Experiments performed on Mayo Clinic test samples for CT image reconstruction from subsampled data. Comparison of mean execution time (in seconds) and the maximum number of performed iterations (if any), for the considered solvers.}
    \label{tab:Mayo_times}
\end{table}

\begin{figure}
    \centering 
    \renewcommand{\arraystretch}{0.5} 
    {\footnotesize
    \begin{tabular}{@{}c c@{\hskip 0.5mm}c c@{\hskip 0.5mm}c}
      & \multicolumn{2}{c}{Training to $\x^{GT}$} & \multicolumn{2}{c}{Training to $\bar{\x}^{(h)}$ }   \\ 
     incTpV & incNN & incDG & incNN & incDG \\ 
    \midrule \rule{0pt}{5mm}  
        RE = 0.0488 & 
        RE = 0.0612 & 
        RE = 0.0585 & 
        RE = 0.0651 & 
        RE = 0.0598 \\ 
        SSIM = 0.9548 & 
        SSIM = 0.9254 & 
        SSIM = 0.9375 & 
        SSIM = 0.9193 & 
        SSIM = 0.9360 \\ 
        \includegraphics[clip, trim = {23mm 17mm 17mm 23mm}, width=0.19\textwidth]{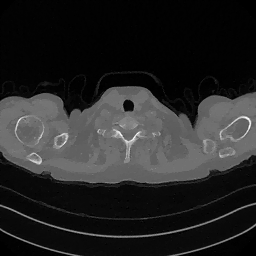} &  
        \includegraphics[clip, trim = {23mm 17mm 17mm 23mm}, width=0.19\textwidth]{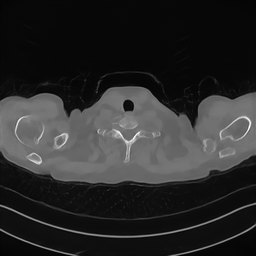} &  
        \includegraphics[clip, trim = {23mm 17mm 17mm 23mm}, width=0.19\textwidth]{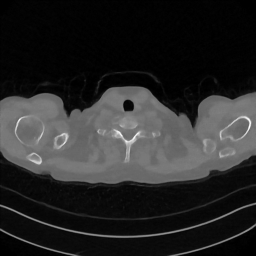} & 
        \includegraphics[clip, trim = {23mm 17mm 17mm 23mm}, width=0.19\textwidth]{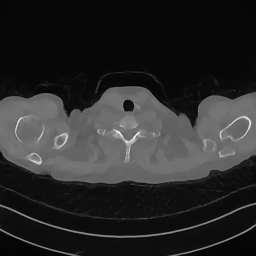} & 
        \includegraphics[clip, trim = {23mm 17mm 17mm 23mm}, width=0.19\textwidth]{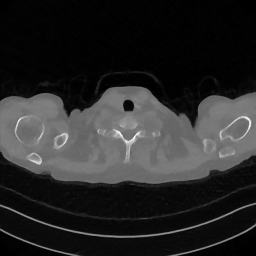}  \\ \rule{0pt}{5mm} 
        RE = 0.0666 & 
        RE = 0.0655 & 
        RE = 0.0676 & 
        RE = 0.0722 & 
        RE = 0.0712 \\ 
        SSIM = 0.9333 & 
        SSIM = 0.9204 & 
        SSIM = 0.9269 & 
        SSIM = 0.9103 & 
        SSIM = 0.9214 \\ 
        \includegraphics[clip, trim = {20mm 25mm 20mm 15mm}, width=0.19\textwidth]{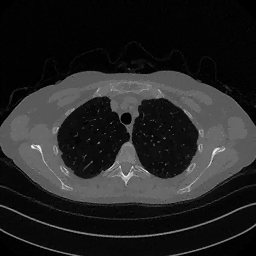} &  
        \includegraphics[clip, trim = {20mm 25mm 20mm 15mm}, width=0.19\textwidth]{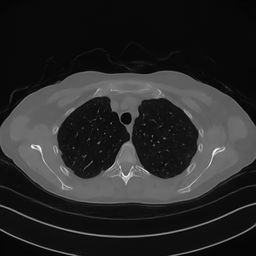} &  
        \includegraphics[clip, trim = {20mm 25mm 20mm 15mm}, width=0.19\textwidth]{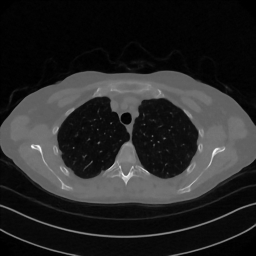} & 
        \includegraphics[clip, trim = {20mm 25mm 20mm 15mm}, width=0.19\textwidth]{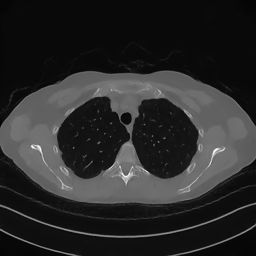} & 
        \includegraphics[clip, trim = {20mm 25mm 20mm 15mm}, width=0.19\textwidth]{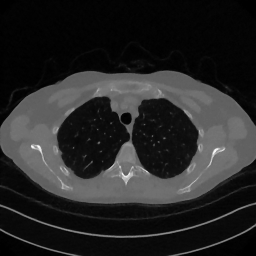} \\ \rule{0pt}{5mm}
        RE = 0.0896 & 
        RE = 0.1875 & 
        RE = 0.1020 & 
        RE = 0.1833 & 
        RE = 0.0974 \\ 
        SSIM = 0.9500 & 
        SSIM = 0.7236 & 
        SSIM = 0.9387 & 
        SSIM = 0.6793 & 
        SSIM = 0.9390 \\ 
        \includegraphics[clip, trim = {35mm 30mm 5mm 10mm}, width=0.19\textwidth]{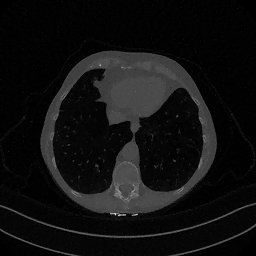} &  
        \includegraphics[clip, trim = {35mm 30mm 5mm 10mm}, width=0.19\textwidth]{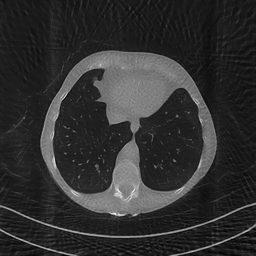} &  
        \includegraphics[clip, trim = {35mm 30mm 5mm 10mm}, width=0.19\textwidth]{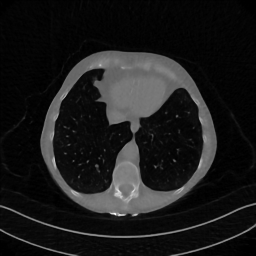} & 
        \includegraphics[clip, trim = {35mm 30mm 5mm 10mm}, width=0.19\textwidth]{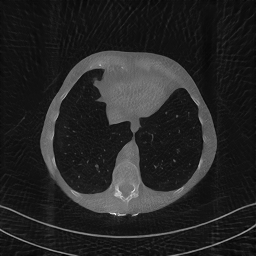} & 
        \includegraphics[clip, trim = {35mm 30mm 5mm 10mm}, width=0.19\textwidth]{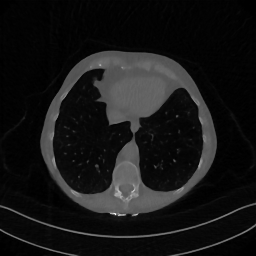}  \\ \rule{0pt}{5mm} 
        RE = 0.0499 & 
        RE = 0.0560 & 
        RE = 0.0496 & 
        RE = 0.0569 & 
        RE = 0.0511 \\ 
        SSIM = 0.9350 & 
        SSIM = 0.9017 & 
        SSIM = 0.9290 & 
        SSIM = 0.9051 & 
        SSIM = 0.9272 \\ 
        \includegraphics[clip, trim = {28mm 28mm 12mm 12mm}, width=0.19\textwidth]{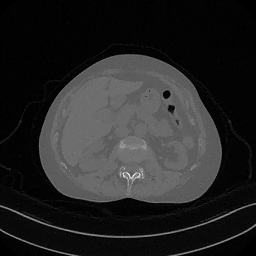} &  
        \includegraphics[clip, trim = {28mm 28mm 12mm 12mm}, width=0.19\textwidth]{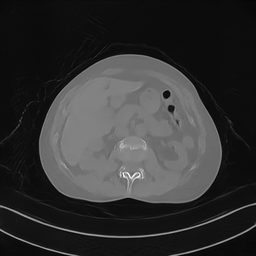} &  
        \includegraphics[clip, trim = {28mm 28mm 12mm 12mm}, width=0.19\textwidth]{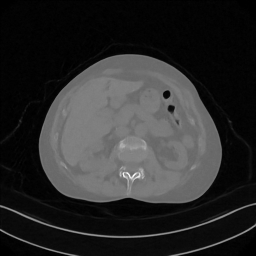} & 
        \includegraphics[clip, trim = {28mm 28mm 12mm 12mm}, width=0.19\textwidth]{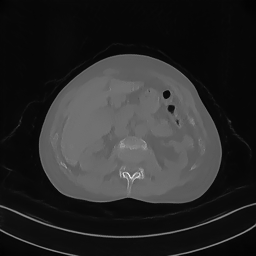} & 
        \includegraphics[clip, trim = {28mm 28mm 12mm 12mm}, width=0.19\textwidth]{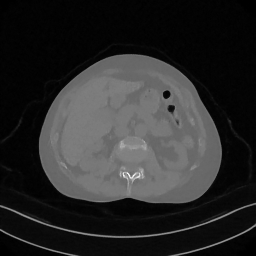} \\     
    \end{tabular}
    } 
    \caption{Experiments performed on Mayo Clinic test samples for CT image reconstruction from subsampled data. Zoomed areas on the solution images from the ground truth ones depicted in Figure \ref{fig:ct_mayoGT}. From left to right: final solution by the inc$\TpV$, by the incNN and the proposed incDG where the neural networks are trained with ground truth images as targets, and by the incNN and the proposed incDG using the incremental solutions $\bar{\x}^{(h)}$ by inc$\TpV$ as targets. }
    \label{fig:ct_mayo}
\end{figure}

\begin{figure}
    \centering
    \includegraphics[width=0.9\linewidth]{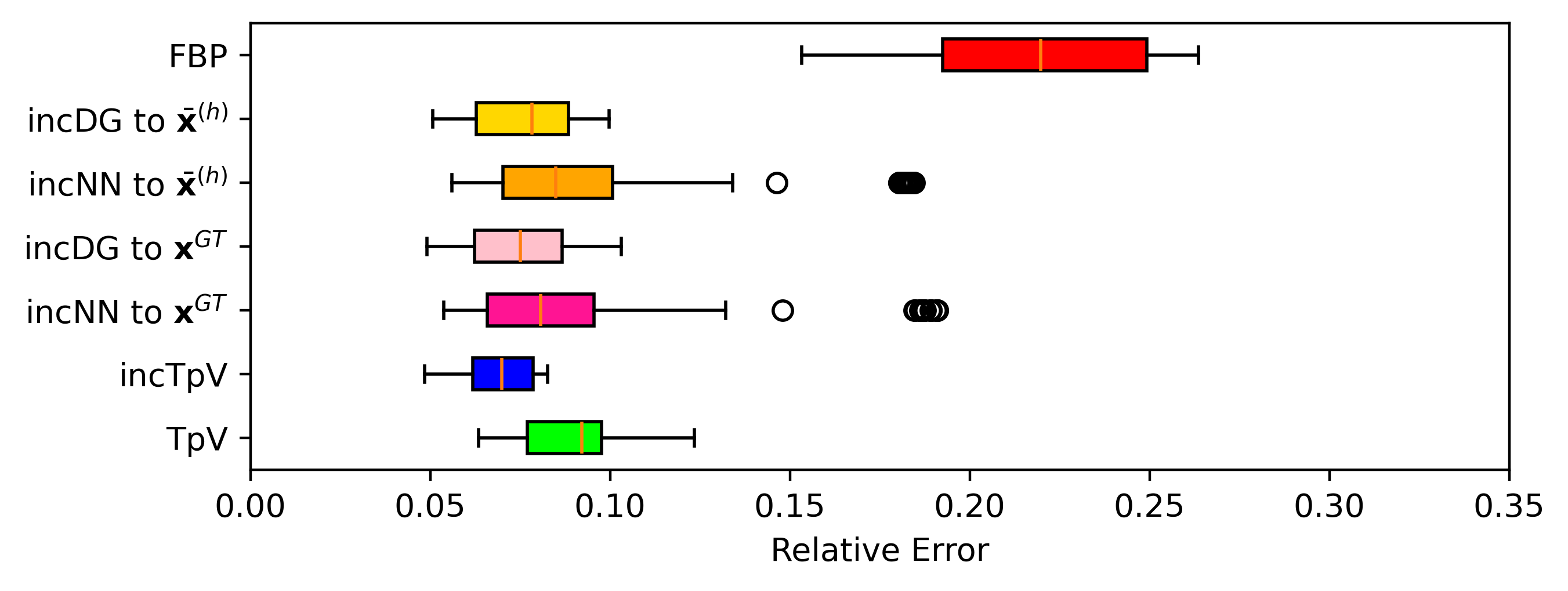} \\ \vspace{2mm}
    \includegraphics[width=0.9\linewidth]{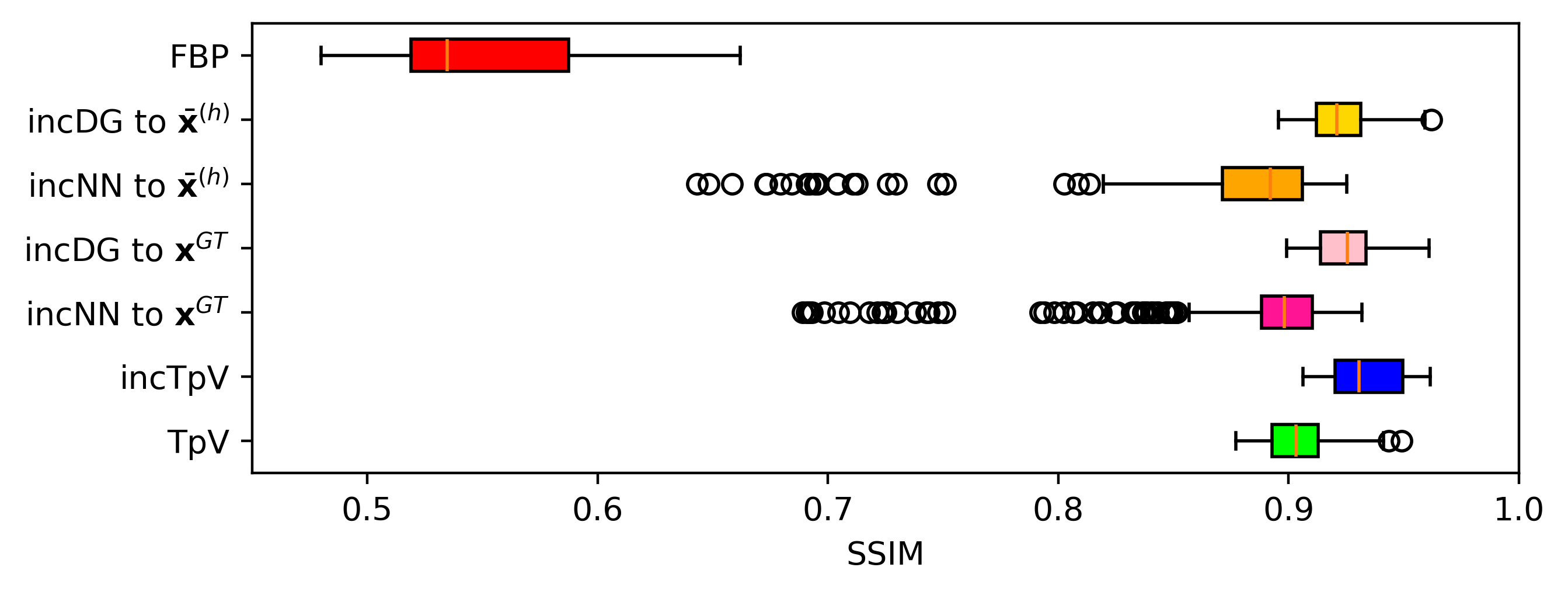}
    \caption{Experiments performed on Mayo Clinic test samples for CT image reconstruction from subsampled data. Boxplot comparisons of the RE metrics (top) and SSIM metrics (bottom), computed over the entire test set. }
    \label{fig:ct_boxplot}
\end{figure}

As previously mentioned, we now examine two slightly different training strategies, depending on the availability or absence of ground truth samples, and the incNN and incDG frameworks are presented in both settings in the following. In Figures \ref{fig:ct_mayo} and \ref{fig:ct_boxplot}, the suffix `to $\x^{GT}$' indicates training where ground truth images are used as targets, whereas `to $\x^{(h)}$' corresponds to training where MB solutions replace the unavailable ground truth, as in Equations \eqref{eq:Psi_0_toIS} and \eqref{eq:Psi_h_toIS}.

From the images in Figure \ref{fig:ct_mayo},  we strongly appreciate that the quality achieved with ground-truth-free training remains high and is fully comparable to that of ground-truth-based training. The reconstructions appear highly similar and yield comparable metrics, reinforcing the practical viability of the proposed incDG approach.

Additionally, we remark that most of the solutions exhibit high-quality reconstructions, with clearly visible fine details in the lungs, and well-delineated low-contrast tissues. 
The inc$\TpV$ outperforms the fast methods, achieving the best RE and SSIM values on the displayed images. 
The (blue) boxplots further confirm its robustness across the entire test set. 

However, the incDG reconstructions are also highly accurate, though slightly affected by minor noise and mild oversmoothing. We also emphasize that incDG consistently outperforms incNN in terms of RE and SSIM values, regardless of the training configuration. For instance, in incDG images, the aorta is more distinctly visible with its segmented tubular structure, and the organ boundaries in the abdomen better resemble those in the reference images.
The benefits of incorporating MB iterations are particularly evident when incNN displays its characteristic instability, as in the third test case. Here, both versions of incNN introduce hallucinations, as artifacts reminiscent of streaking patterns commonly observed in CT reconstructions from undersampled data. They are clearly visible in the background, and they distort (or even remove) significant anatomical structures. Fortunately, these artifacts are effectively mitigated in the incDG solutions, well preserving image features.
The unstable behaviour of incNN is also evident in its (pink and orange) boxplots, reflecting the presence of many outliers with poor SSIM, whereas the incDG (light-pink and yellow boxplots)  exhibits the stable trend typical of mathematically-grounded algorithms.

\section{Conclusions}\label{sec:concl}
This study has first designed the inc$\TpV$ scheme, a model-based incremental algorithm specifically tailored for total $p$-variation ($\TpV$) regularized imaging problems. The approach progressively enhances sparsity in the gradient domain, exploiting the iterative reweighted strategy to handle the non-convex regularization. Building on this, we introduced incDG, a hybrid framework that integrates pretrained neural networks to accelerate the incremental optimization process. Thus, the proposed incDG approach effectively addresses the non-convex inverse problem, combining the robustness of variational methods with the efficiency of deep learning to iteratively approach the optimal $\ell_0$ solution.
The incDG strategy effectively integrates the advantages of deep learning and model-based optimization. On one hand, the Deep Guess mechanism, based on a neural network, provides high-quality initializations for non-convex minimization. On the other, the iterative solver maintains the robustness and flexibility of model-based regularized optimization, ensuring reliability in diverse scenarios.

Specifically conceived for medical image processing, incDG aligns with the needs of clinical software, which prioritizes fast image enhancement and accurate segmentation to support medical decisions. Additionally, the inherent anatomical variability in real-world medical imaging poses a challenge for commercial algorithms, further highlighting the importance of a robust framework.

Extensive numerical results confirm the flexibility and efficiency of incDG in addressing $\TpV$ regularization for image deblurring and reconstruction tasks. The method has demonstrated strong performance across a range of human CT images, from brain hemorrhages to chest scans. Compared to purely iterative model-based approaches, incDG is faster and, in most cases, achieves superior results. When compared to deep learning-based methods, it exhibits higher accuracy and significantly greater stability, demonstrating robustness to input variations and consistently high performance across different images.

Moreover, when trained in a ground-truth-free manner, incDG maintains high reconstruction quality with minimal degradation. This resilience further enhances its practical usability, making it a compelling tool for medical imaging applications.




\bibliographystyle{elsarticle-num}

\subsubsection*{Acknowledgement}
The author is partially supported by the “Fondo per il Programma Nazionale di Ricerca e Progetti di Rilevante Interesse Nazionale (PRIN)” 2022 project “STILE: Sustainable Tomographic Imaging with Learning and rEgularization”, project code: 20225STXSB, funded by the European Commission under the NextGeneration EU programme, project code MUR 20225STXSB, CUP J53D23003600006.\\
The author  is also partially supported by the Gruppo Nazionale per il Calcolo Scientifico (INdAM - GNCS) within the projects `Metodi avanzati di ottimizzazione stocastica per la risoluzione di problemi inversi di imaging', project code CUP$\_$E53C24001950001.

\end{document}